\documentclass{article}

\usepackage{microtype}
\usepackage{graphicx}
\usepackage{subcaption}
\usepackage{booktabs} % for professional tables
\usepackage{multirow}
\usepackage{array}
\usepackage{hyperref}

\usepackage[preprint]{arXiv}

\makeatletter
\renewcommand{\ICML@preprint}{} % (or put your own text here)
\makeatother

\usepackage{amsmath}
\usepackage{amssymb}
\usepackage{mathtools}
\usepackage{amsthm}

\usepackage[capitalize,noabbrev]{cleveref}

\theoremstyle{plain}

\theoremstyle{definition}

\theoremstyle{remark}

\usepackage[disable,textsize=tiny]{todonotes}
\newcommand{\stephead}[1]{\par\noindent\textbf{#1.}\ }
\newcommand{\substep}[1]{\noindent\textit{#1}\ }
\newcommand{\reshead}[1]{\par\noindent\textbf{#1}\ }
\usepackage{adjustbox}
\newcolumntype{Y}{>{\raggedright\arraybackslash}X}

\icmltitlerunning{Do Generative Metrics Predict YOLO Performance?}

\begin{document}

\twocolumn[
  \icmltitle{Do Generative Metrics Predict YOLO Performance? An Evaluation Across Models, Augmentation Ratios, and Dataset Complexity}

   % List of affiliations: The first argument should be a (short) identifier you
  % will use later to specify author affiliations Academic affiliations
  % should list Department, University, City, Region, Country Industry
  % affiliations should list Company, City, Region, Country

  % You can specify symbols, otherwise they are numbered in order. Ideally, you
  % should not use this facility. Affiliations will be numbered in order of
  % appearance and this is the preferred way.
  \icmlsetsymbol{equal}{*}
  \begin{icmlauthorlist}
    \icmlauthor{Vasile Marian}{uq}
    \icmlauthor{Yong-Bin Kang}{sw}
    \icmlauthor{Alexander Buddery}{uq}
    %\icmlauthor{Firstname4 Lastname4}{sch}
    %\icmlauthor{Firstname5 Lastname5}{yyy}
    %\icmlauthor{Firstname6 Lastname6}{sch,yyy,comp}
    %\icmlauthor{Firstname7 Lastname7}{comp}
    %\icmlauthor{}{sch}
    %\icmlauthor{Firstname8 Lastname8}{sch}
    %\icmlauthor{Firstname8 Lastname8}{yyy,comp}
    %\icmlauthor{}{sch}
    %\icmlauthor{}{sch}
  \end{icmlauthorlist}

  \icmlaffiliation{uq}{School of Mechanical and Mining Engineering, The University of Queensland,Level 4, Mansergh Shaw Building (45), St Lucia, QLD 4072, Australia}
  \icmlaffiliation{sw}{Swinburne University of Technology, John Street, Hawthorn, VIC 3122, Australia}
  
  \icmlcorrespondingauthor{Vasile Marian}{vasile.marian@student.uq.edu.au}

 % You may provide any keywords that you find helpful for describing your
  % paper; these are used to populate the "keywords" metadata in the PDF but
  % will not be shown in the document
  \icmlkeywords{Object Detection, Synthetic Data, Generative Models, Dataset Metrics, YOLO}

  \vskip 0.3in
]

% this must go after the closing bracket ] following \twocolumn[ ...

% This command actually creates the footnote in the first column listing the
% affiliations and the copyright notice. The command takes one argument, which
% is text to display at the start of the footnote. The \icmlEqualContribution
% command is standard text for equal contribution. Remove it (just {}) if you
% do not need this facility.

% Use ONE of the following lines. DO NOT remove the command.
% If you have no special notice, KEEP empty braces:
\printAffiliationsAndNotice{}  % no special notice (required even if empty)
% Or, if applicable, use the standard equal contribution text:
% \printAffiliationsAndNotice{\icmlEqualContribution}

\begin{abstract}
  Synthetic images are increasingly used to augment object-detection training sets, but reliably evaluating a synthetic dataset before training remains difficult: standard global generative metrics (e.g., FID) often do not predict downstream detection mAP. We present a controlled evaluation of synthetic augmentation for YOLOv11 across three single-class detection regimes—Traffic Signs (sparse/near-saturated), Cityscapes Pedestrian (dense/occlusion-heavy), and COCO PottedPlant (multi-instance/high-variability). We benchmark six GAN-, diffusion-, and hybrid-based generators over augmentation ratios from 10\% to 150\% of the real training split, and train YOLOv11 both from scratch and with COCO-pretrained initialization, evaluating on held-out real test splits (mAP@0.50:0.95). For each dataset--generator--augmentation configuration, we compute pre-training dataset metrics under a matched-size bootstrap protocol, including (i) global feature-space metrics in both Inception-v3 and DINOv2 embeddings and (ii) object-centric distribution distances over bounding-box statistics. Synthetic augmentation yields substantial gains in the more challenging regimes (up to +7.6\% and +30.6\% relative mAP in Pedestrian and PottedPlant, respectively) but is marginal in Traffic Signs and under pretrained fine-tuning. To separate metric signal from augmentation quantity, we report both raw and augmentation-controlled (residualized) correlations with multiple-testing correction, showing that metric–performance alignment is strongly regime-dependent and that many apparent raw associations weaken after controlling for augmentation level.
\end{abstract}
\section{Introduction}
\label{sec:intro}

Data augmentation is a standard tool for improving object detection, especially when labeled training data are limited, imbalanced, or expensive to collect \citep{Sun2017Revisiting, Shao2019Objects365, Peng2020LargeScaleDetection, Michaelis2020Generalization}. While scaling up real, annotated datasets can yield strong gains, it is often costly or impractical in specialized domains and under privacy or access constraints \citep{Jaipuria2020DatasetBias, zoph2019learningdataaugmentationstrategies, Sun2017Revisiting}, motivating \emph{synthetic images} produced by simulation or modern generative models as an additional source of training data.

Recent GANs and diffusion models can generate high-fidelity images across many domains \citep{c:goodfellow2014generative, ho2020denoisingdiffusionprobabilisticmodels}, and synthetic augmentation has been explored in both general and niche detection settings \citep{Tobin2017, Tremblay2018, Barbieri_2021_WACV, Fescenko2023Bottles}. However, a practical bottleneck remains: how can we evaluate and prioritize synthetic data sources \textbf{before} training a detector? In real pipelines, multiple candidate generators and augmentation ratios are available, but generating, labeling, and verifying synthetic data at scale is expensive; practitioners therefore need principled \emph{screening signals} to decide which generator to prioritize and how much synthetic data to produce. We use \emph{usefulness} to mean \textbf{downstream detection benefit}, i.e., improved COCO-style mAP@0.50:0.95 on a held-out \emph{real} test set after augmenting training with synthetic data.

A common approach is to rank synthetic datasets using global, image-level generative metrics such as Inception Score (IS) \citep{Salimans2016} and Fr\'echet Inception Distance (FID) \citep{Heusel2017}, including related precision/recall and coverage-style variants computed in Inception-v3 feature space. Yet multiple studies report that similar FID values can correspond to substantially different detector performance, and that improvements in standard generative metrics do not consistently translate into higher detection mAP \citep{Lin2023FewShotSynthetic, Zhou2025SyntheticDatasetGeneration, Zenith2025SDQM}. Thus, it remains unclear which \emph{pre-training} dataset metrics, if any, reliably predict whether synthetic augmentation from a given generator will improve YOLO-style detection performance.

Three factors make this gap particularly challenging for object detection. First, global metric behavior depends on the representation used: Inception-v3 features are standard for FID/IS, but their suitability can vary across domains and modern generative outputs, motivating comparisons with alternative self-supervised encoders such as DINOv2 \citep{Stein2023ExposingFlaws, Oquab2023DINOv2}. Second, detector training depends on object-level structure (instance density, scale distributions, and difficulty or occlusion patterns) that may not be well summarized by global realism or diversity scores; motivated by dataset-shift and domain-gap analysis, simple distribution distances over label-derived statistics provide a lightweight way to quantify object-centric mismatch \citep{Doan2024DomainGap, Xu2019WassersteinDA, Xu2022NWD}. Third, augmentation effectiveness is regime-dependent: synthetic data may help in dense, small-object, occlusion-heavy scenes but saturate or have negligible effect in simpler regimes.

To study these issues under a controlled pipeline, we evaluate three single-class detection datasets spanning distinct regimes: \textbf{Traffic Signs} \citep{trafficSignsInternal2026} (sparser, low-overlap scenes), \textbf{Cityscapes Pedestrian} \citep{Cordts2016Cityscapes} (dense, occlusion-heavy scenes with many small objects), and \textbf{COCO PottedPlant} \citep{Lin2014COCO} (multi-instance scenes with wide scale variation and diverse backgrounds). We use YOLOv11 \citep{yolo11_ultralytics} as the downstream detector and evaluate both \textbf{from scratch} training and \textbf{COCO-pretrained} fine-tuning, since initialization can change sensitivity to synthetic data and may be unavailable or unreliable under strong domain shift or deployment constraints. 

\noindent\textbf{This leads to three research questions:}
\textbf{(Q1)} Do global generative metrics track YOLOv11 performance, and how does this depend on the encoder used to compute them (Inception-v3 vs.\ DINOv2)?
\textbf{(Q2)} Do simple object-centric distribution metrics (e.g., Wasserstein or Jensen--Shannon distances over bounding-box statistics) provide complementary signal for explaining YOLOv11 outcomes?
\textbf{(Q3)} How do metric--performance relationships vary across dataset regime and initialization (from scratch vs.\ pretrained), and can metrics support practical generator prioritization under a fixed synthetic-data budget?

\noindent\textbf{This work.} We conduct a controlled evaluation of synthetic augmentation and metric--performance alignment for YOLOv11 across multiple generators and augmentation ratios (10\%--150\% relative to the real training split). For each dataset--generator--augmentation configuration, we compute \emph{pre-training} synthetic-dataset metrics from two families: global encoder-based metrics computed in both Inception-v3 and DINOv2 feature spaces, and object-centric distribution metrics computed from bounding-box statistics. We then train YOLOv11 in both from scratch and pretrained regimes and evaluate on held-out \emph{real} test splits. To isolate metric signal relevant to fixed-budget screening, we analyze both raw and augmentation-controlled (residualized) correlations with multiple-testing control.

\noindent\textbf{Contributions:}
\textbf{(i)} A systematic benchmark of YOLOv11 synthetic augmentation across three detection regimes, six generators, seven augmentation ratios (10\%--150\%), and two initialization regimes (from scratch vs.\ COCO-pretrained);
\textbf{(ii)} A comparative analysis of global generative metrics computed with Inception-v3 versus DINOv2 features, alongside object-centric distribution metrics based on bounding-box statistics, characterizing when each aligns with detector mAP;
\textbf{(iii)} An evaluation protocol that controls for augmentation ratio when assessing metric--performance alignment, supporting exploratory fixed-budget generator prioritization rather than correlations dominated by augmentation amount.

\section{Related Work}
\label{sec:related-work}

We review work most relevant to our study: (i) YOLO-based object detection, (ii) traditional and generative augmentation for detection, and (iii) whether \emph{pre-training} synthetic-data metrics predict downstream detector performance.

\noindent \textbf{YOLO-based object detection.}
YOLO-style detectors reformulate detection as a single-stage prediction problem, enabling real-time inference with strong accuracy--speed trade-offs and broad adoption across application domains \citep{Redmon2016, redmon2018yolov3, Bochkovskiy2020, Jocher2023, yolo11_ultralytics}. Performance gains across versions stem from both architectural advances and training ``bag-of-tricks'' (including augmentation and optimization improvements) \citep{Bochkovskiy2020, wang2022yolov7trainablebagoffreebiessets}. While large benchmarks such as MS-COCO, Open Images, and Cityscapes have driven progress \citep{Lin2014COCO, Kuznetsova_2020, Cordts2016Cityscapes}, performance often degrades in limited-label regimes, where augmentation and transfer learning remain practical tools \citep{Sun2017Revisiting, zoph2019learningdataaugmentationstrategies, cui2023generalizedfewshotcontinuallearning, wang2020generalizingexamplessurveyfewshot}.

\noindent \textbf{Data augmentation for detection.}
Classic augmentations improve generalization by transforming existing images, including geometric and photometric transforms, occlusion-based methods, and compositional strategies such as MixUp, CutMix, and Mosaic \citep{Shorten2019SurveyAug, perez2017effectivenessdataaugmentationimage, NIPS2012_c399862d, devries2017improvedregularizationconvolutionalneural, chen2024gridmaskdataaugmentation, zhang2018mixupempiricalriskminimization, yun2019cutmixregularizationstrategytrain, Bochkovskiy2020}. These methods are inexpensive and widely used in modern YOLO pipelines, but they cannot reliably introduce \emph{new} object instances, layouts, or backgrounds, and thus may saturate when missing modes limit detection \citep{He2023DGGAN, zhu2024odgendomainspecificobjectdetection}.
\begin{figure*}[t]
\centering
\includegraphics[width=1\linewidth]{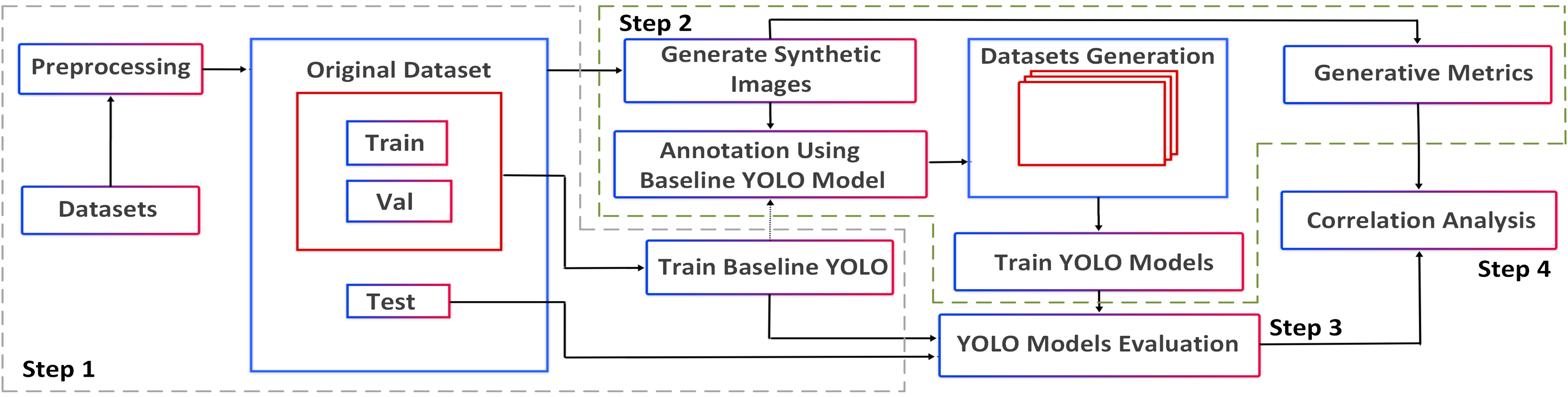}
\caption{\small{Overview of the four-stage pipeline: dataset curation \& baselines; synthetic generation \& labeling; YOLO training/evaluation; metric--performance analysis.}}
\label{fig:Sec3_Diag}
\end{figure*}
\noindent \textbf{Generative synthetic augmentation.}
GANs and diffusion models provide a route to generating additional training images beyond transformations of existing data \citep{c:goodfellow2014generative, ho2020denoisingdiffusionprobabilisticmodels}. GAN-based augmentation has been used in several detection domains, but visual realism alone does not guarantee detector gains because viewpoint diversity, label consistency, and occlusion patterns are often decisive \citep{bay2025impactsyntheticdataobject, guo2025neptune-x, han2018improvingfacedetectionperformance}. Diffusion-based pipelines have recently enabled controllable synthesis, inpainting-based variation, and annotation reuse, with growing interest in detector-aware generation objectives \citep{fang2024data, zhu2024odgendomainspecificobjectdetection, Wang2024DetDiffusion}. Despite these advances, the field still lacks consensus on when synthetic data helps, how much to add, and which dataset properties drive improvements.

\noindent \textbf{Do generative metrics predict detector performance?}
Many studies evaluate synthetic augmentation using downstream mAP, but results are mixed: higher-fidelity samples can improve YOLO-family detectors in some settings, while other work reports diminishing returns or failures when synthetic data lacks domain-relevant diversity or induces distribution shift \citep{Tremblay2018TrainingDeepSynthetic, electronics13071231, He2023DGGAN, 9471877, modak2024enhancingweeddetectionperformance, app15010354, 10.1117/12.3015657, xiang2024dodadiffusionobjectdetectiondomain, s21237901, Dwibedi2017CutPaste, Kar2019MetaSim}. Correspondingly, common global image-level metrics such as IS and FID, while useful for measuring perceptual realism and diversity, often correlate weakly or inconsistently with detector mAP \citep{Salimans2016, Heusel2017}. Recent work has emphasized limitations of FID under modern generative settings and proposed alternatives using richer representations \citep{Jayasumana_2024_CVPR}. There is also increasing interest in \emph{detection-oriented} dataset metrics that reflect object-level properties (e.g., coverage, alignment, occlusion) and can better explain downstream behavior \citep{zhu2024odgendomainspecificobjectdetection, kynkäänniemi2019improvedprecisionrecallmetric, Zenith2025SDQM}. However, systematic evidence on how global metrics (across encoders) and simple object-centric distribution metrics align with YOLO performance across augmentation ratios and dataset regimes remains limited. Our work addresses this by benchmarking multiple metric families and explicitly controlling for augmentation amount when assessing metric--performance alignment.

\section{Evaluation Framework: Synthetic Augmentation and Metric--Performance Analysis for YOLOv11}
\label{sec:methodology}

To study which properties of synthetic data are informative about downstream detector behavior, we define a structured evaluation protocol (Figure~\ref{fig:Sec3_Diag}) that connects (i) synthetic data generation and dataset construction, (ii) pre-training dataset metric computation, and (iii) YOLOv11 training and evaluation under controlled experimental factors (dataset, generator family, augmentation ratio, and initialization regime).

Our evaluation follows a four-stage pipeline (Figure~\ref{fig:Sec3_Diag}) over a controlled
dataset$\times$generator$\times$augmentation grid. \textbf{Step 1:} curate and preprocess three single-class detection datasets and train real-only YOLOv11 baselines under both from scratch and COCO-pretrained initialization. \textbf{Step 2:} generate synthetic images, label them with a teacher-assisted, human-audited workflow, build augmented training sets (10\%--150\%), and compute pre-training dataset metrics using a matched-size bootstrap for comparability across ratios; metrics include global embedding-space scores in Inception-v3 and DINOv2 features and object-centric distances over bounding-box statistics. \textbf{Step 3:} evaluate the YOLOv11 checkpoints produced in Step~2 on held-out \emph{real} test splits (mAP@0.50:0.95) with fixed settings. \textbf{Step 4:} quantify metric--performance alignment via raw and augmentation-controlled (residualized) correlations with multiple-testing correction, targeting fixed-budget generator screening.  We next detail Steps~1--4, operationalizing Q1--Q3 under a controlled dataset $\times$ generator $\times$ augmentation-ratio grid and two initialization regimes. For clarity, validation/test splits remain real-only and all training settings are held fixed across comparisons.

\begin{figure}[!h]
\centering
\includegraphics[width=\linewidth]{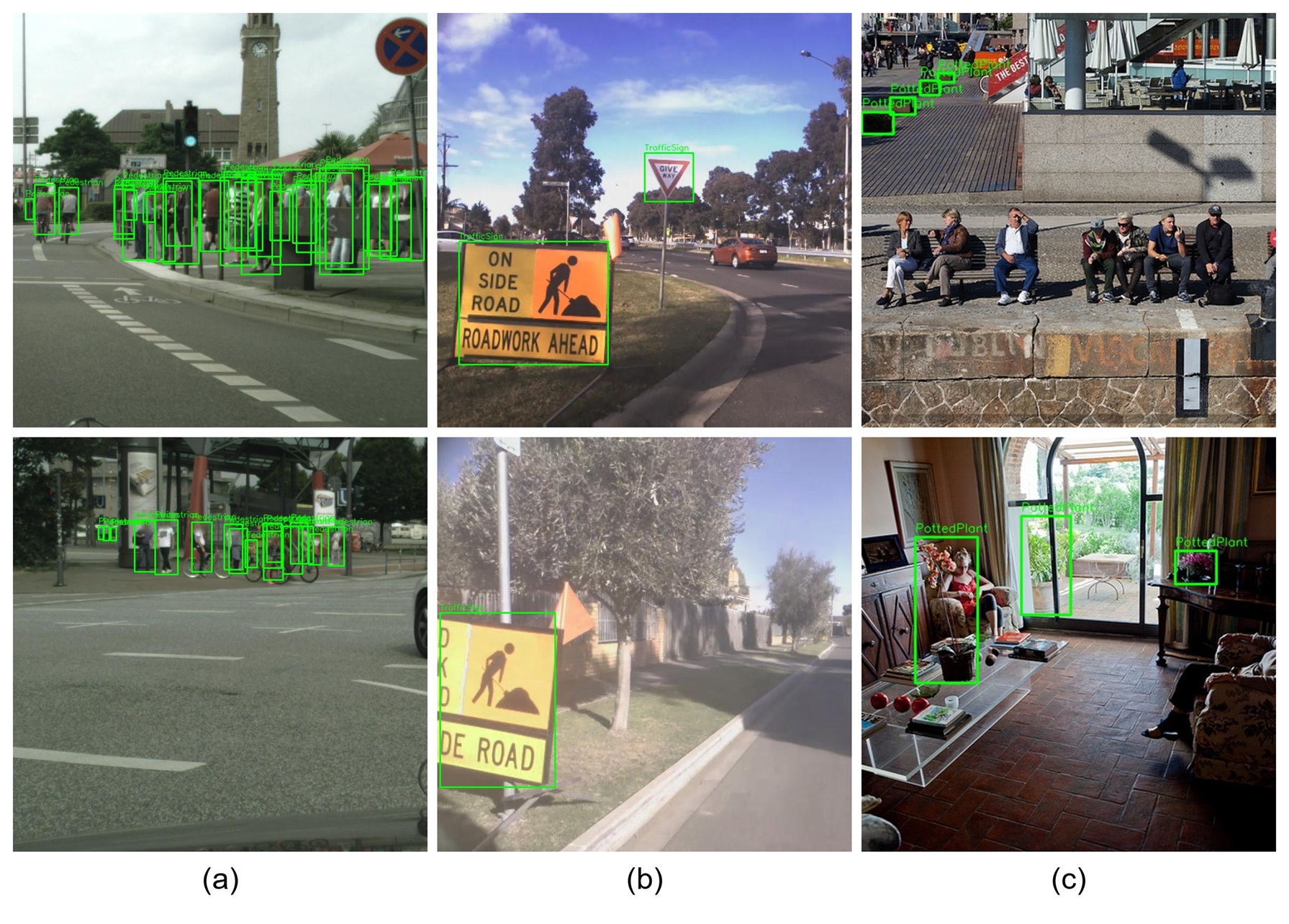}
\caption{Representative preprocessed samples from the three datasets used in this study: (a) Cityscapes Pedestrian, (b) Traffic Signs, and (c) COCO PottedPlant.}
\label{fig:SamplesGrid}
\end{figure}

\stephead{Step 1: Datasets, preprocessing, and real-only baselines}
We evaluate three single-class detection datasets chosen to span distinct detection regimes (Table~\ref{tab:dataset_summary}).
\begin{table}[h]
\centering
\caption{\small Summary of the three evaluation datasets after preprocessing and curation. All experiments use an approximately 70/15/15 train/val/test split; full curation details and additional examples are in Appendix~\ref{sec:supp-dataset-samples}.}
\label{tab:dataset_summary}
\resizebox{1.00\columnwidth}{!}{%
\begin{tabular}{l c l c l}
\toprule
\textbf{Dataset} & \textbf{\# imgs} & \textbf{Target class mapping} & \textbf{Final size} & \textbf{Regime tag} \\
\midrule
Cityscapes Pedestrian & 2195 & person+rider+sitting $\rightarrow$ pedestrian & $768{\times}768$ & dense / occlusion-heavy \\
Traffic Signs & 2568 & all sign classes $\rightarrow$ traffic sign & $768{\times}768$ & sparse / near-saturated \\
COCO PottedPlant & 2568 & potted plant $\rightarrow$ potted plant & $512{\times}512$ & multi-instance / diverse \\
\bottomrule
\end{tabular}
}
\end{table}
Figure~\ref{fig:SamplesGrid} shows representative preprocessed samples.
Additional real examples and label-derived regime statistics are reported in Appendix~\ref{sec:supp-dataset-samples}
(Appendix~\ref{app:dataset_examples}, Table~\ref{tab:dataset_regime_stats}).
Dataset curation and preprocessing details are provided in Appendix~\ref{app:dataset_curation_details}.

We split each dataset into train/validation/test using an approximately $70\%/15\%/15\%$ split.
Validation and test splits remain real-only throughout.
We train real-only YOLOv11 baselines for each dataset under from scratch and COCO-pretrained initialization.
All augmented runs reuse the same training recipe and checkpoint-selection procedure.
Training settings are summarized in Appendix~\ref{app:yolo_training_details}. We use Ultralytics deterministic training with a fixed seed (\texttt{seed}=0) for comparability across the full grid, and we report a targeted 3-seed robustness check on representative baseline/augmented comparisons in Appendix~\ref{app:seed_robustness}.

\stephead{Step 2: Synthetic data generation, labeling, and metric computation}
\label{sec:step2}
\substep{Synthetic generation, annotation, and augmented training sets.} For each dataset, we generate \textbf{3{,}000} synthetic images per generator at $256{\times}256$ resolution using six models spanning diffusion, GAN, and hybrid families: DiT~\citep{c:peebles2023scalable}, ADM~\citep{c:nichol2021improveddenoisingdiffusionprobabilistic}, DiffusionGAN~\citep{c:wang2023diffusiongan}, StyleGAN2-ADA~\citep{c:karras2020training}, ProjectedGAN~\citep{c:sauer2021projected}, and LayoutDiffusion~\citep{Zheng_2023_CVPR}. Because synthetic images lack ground-truth boxes, we adopt a \emph{model-assisted} annotation workflow: for each dataset, a fixed reference YOLOv11 detector (COCO-pretrained and fine-tuned on that dataset's \emph{real} training split) generates initial candidate boxes using the same thresholds for all generators (Appendix~\ref{app:step2_impl_details}). We then manually audit and correct the generated annotations (removing false positives, fixing boxes, and adding missed instances when present), and we remove synthetic images that contain no target object after inspection (single-class setting) to avoid introducing unlabeled positives as background. This ``detector proposes, human corrects'' protocol follows standard semi-automatic bounding-box annotation practice used to reduce labeling cost while maintaining label quality~\citep{Adhikari2020IterativeBBoxAnnotation, Konyushkova2017LearningDialogsBBox, Liu2021UnbiasedTeacher, Xu2021SoftTeacher}.

Using the resulting labeled synthetic pool, we construct augmented training sets by adding synthetic images to the real training split at ratios of 10\%, 25\%, 50\%, 75\%, 100\%, 125\%, and 150\%. For each dataset--generator--augmentation configuration, we train YOLOv11 under two initialization regimes (random initialization and COCO-pretrained fine-tuning). Validation and test splits remain real-only so that reported performance reflects generalization to real data. All training runs use identical optimizer settings, schedule, batch size, image size, and early-stopping criteria; only training-set composition and initialization differ. We set \texttt{imgsz}=768 for Cityscapes Pedestrian and Traffic Signs and \texttt{imgsz}=512 for COCO PottedPlant (Ultralytics letterbox resize); synthetic $256{\times}256$ images are resized through the same pipeline.

\substep{Pre-training metric families and matched-size protocol.} Before training YOLO, we compute dataset-level metrics for each dataset--generator--augmentation configuration from three families: (i) global embedding-space metrics in Inception-v3 features (e.g., FID and precision/recall and density/coverage-style scores)~\citep{Heusel2017, Sajjadi2018, Naeem2020} using InceptionV3~\citep{Szegedy2015}; (ii) global embedding-space metrics in DINOv2 features~\citep{Oquab2023DINOv2}; and (iii) object-centric distribution distances over bounding-box statistics (Wasserstein/Jensen--Shannon)~\citep{Villani2009, Lin1991}. Because Fr\'echet-style distances are sample-size sensitive~\citep{Chong2020}, embedding-based metrics are computed using a matched-size bootstrap protocol that fixes the real/synthetic subset size per dataset and reports the mean over bootstrap trials. The exact matched-size and bootstrap settings are in Appendix~\ref{app:step2_impl_details}, and full metric definitions are in Appendix~\ref{sec:extended-metrics}.

\stephead{Step 3: YOLO Models Evaluation}
After training, we evaluate each model with the Ultralytics validation framework on the same held-out real test split used for baseline evaluation, using identical settings across runs (including image size and COCO-style mAP@0.50:0.95), so differences reflect only training-set composition and initialization rather than evaluation-time configuration; results are compiled in Section~\ref{sec:results}. During training, Ultralytics reports per-epoch detection metrics on the validation split; we refer to these as \emph{training-time validation} results and, when plotting augmentation curves, use the best validation mAP observed during training for each configuration. Separately, we run Ultralytics \texttt{val} on the unseen real test split for \emph{held-out test evaluation}, which we summarize in Table~\ref{tab:best_map_summary} and plot in Appendix~\ref{app:test_aug_curves}.

\stephead{Step 4: Metric--Performance Correlation Analysis}
In this step, we quantify metric--performance alignment by relating the synthetic-dataset metrics computed in Step~2 to downstream YOLOv11 detection performance from Step~3. We perform the analysis separately for each dataset and initialization regime (from scratch vs.\ COCO-pretrained), so that relationships can be interpreted in the context of dataset characteristics and training regime. Each experimental condition (generator, augmentation ratio) yields one data point consisting of a vector of pre-training metric values and a corresponding YOLOv11 mAP evaluated on the held-out real test split.

A key confounder is the augmentation ratio itself, which can be a dominant driver of mAP. We therefore (i) summarize the overall \emph{augmentation--performance} relationship, and (ii) test whether dataset metrics provide association \emph{beyond} augmentation amount. Concretely, for each metric we report both Pearson correlation (linear association) and Spearman rank correlation (monotonic association, robust to non-linear trends). We compute correlations for (a) \emph{global, encoder-based metrics} computed using \textbf{Inception-v3} features, (b) \emph{global, encoder-based metrics} computed using \textbf{DINOv2} features, and (c) \emph{object-centric distribution metrics} computed from bounding-box statistics using Wasserstein/Jensen--Shannon distances (e.g., per-image object-count distributions, small/difficult-object prevalence, and derived per-image ``complexity'' distributions). Full metrics definitions and details are provided in Appendix~\ref{app:metric_definitions}. 

To assess metric--performance alignment beyond the dominant effect of augmentation ratio, we report both raw and augmentation-controlled (residualized) correlations. The residualization procedure is defined in Section~\ref{sec:results-residual-motivation} (with implementation details summarized in Appendix~\ref{app:residual_correlation_details}).

\section{Results}
\label{sec:results}

We report YOLOv11 detection performance (COCO-style mAP@0.50:0.95) under synthetic augmentation across three single-class datasets (Cityscapes Pedestrian, Traffic Signs, and COCO PottedPlant) and two initialization regimes (training from scratch and COCO-pretrained fine-tuning). For each dataset and regime, we evaluate a common grid of dataset--generator--augmentation configurations (six generators, seven augmentation ratios from 10\% to 150\%, plus a real-only baseline). In addition to reporting mAP, we analyze how mAP co-varies with dataset-level quality metrics computed from matched real/synthetic subsets for each configuration (Section~\ref{sec:methodology}).

\subsection{YOLOv11 performance under synthetic augmentation}
\label{sec:results-augmentation}

\begin{figure}[!ht]
    \centering
    \includegraphics[width=0.97\linewidth]{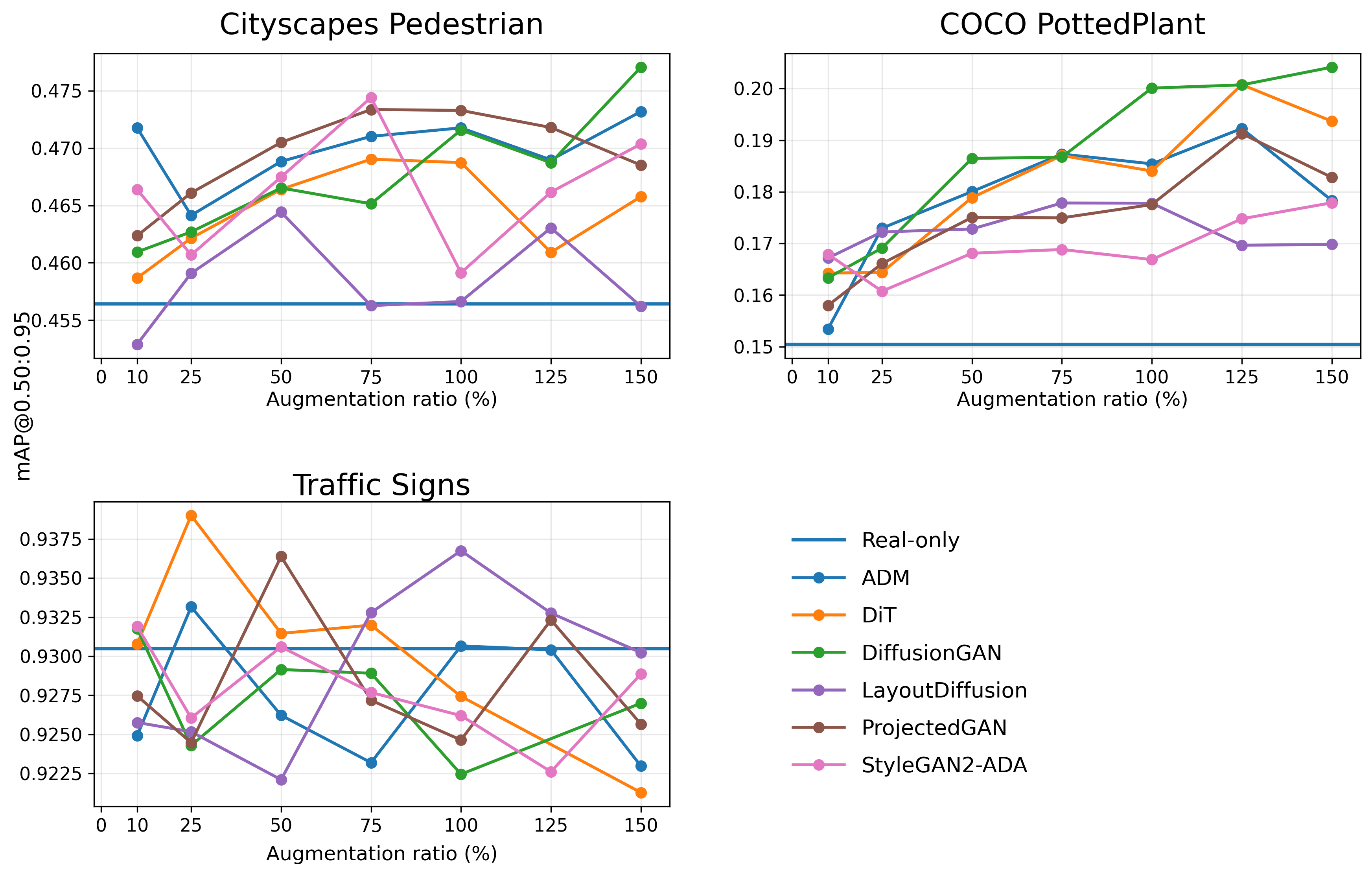}
    \caption{\small YOLOv11 \textbf{From-scratch} \emph{training-time validation} mAP@0.50:0.95 vs.\ augmentation ratio. Each curve corresponds to one generator; the horizontal line indicates the real-only baseline. Curves report the \emph{best validation mAP observed during training} (from Ultralytics training logs).}
    \label{fig:aug_curves_fromscratch}
\end{figure}

\reshead{From-scratch training-time validation curves (checkpoint-selection view)}
\noindent\textbf{Overview.} Figure~\ref{fig:aug_curves_fromscratch} shows \emph{training-time validation} mAP@0.50:0.95 versus augmentation ratio for YOLOv11 trained \textbf{from scratch}. Each point is the \emph{best validation mAP observed during training} for a given generator and augmentation ratio (from Ultralytics training logs). These curves are used to visualize training-time behavior and to select the best checkpoint per run. Pretrained training-time validation curves are provided in Appendix~\ref{app:pretrained_curves}. 
\reshead{From-scratch Held-out test evaluation summary (primary reported results)}
\noindent\textbf{Evaluation protocol.} The primary results we report are \emph{held-out test evaluation} scores: for each training run, we take the checkpoint selected by validation mAP and then run Ultralytics \texttt{val} on the unseen \emph{real} test split. Table~\ref{tab:best_map_summary} summarizes, for each dataset and regime, the real-only baseline versus the best augmented configuration. Full per-generator test-set curves are shown in Appendix~\ref{app:test_aug_curves}.
\begin{table}[t]
\centering
\caption{\small Best \emph{held-out test evaluation} mAP@0.50:0.95 (real-only baseline vs.\ best augmented) for each dataset and initialization regime, including absolute and relative gains. Held-out test results are obtained by running Ultralytics \texttt{val} on the unseen real test split using the best checkpoint from training (selected by validation mAP).}
\label{tab:best_map_summary}

\begingroup
\setlength{\tabcolsep}{4pt}        % default is 6pt; slightly tighter -> less downscaling
\renewcommand{\arraystretch}{1.05} % tiny row breathing room (optional)

\resizebox{\columnwidth}{!}{%
\begin{tabular}{@{}l l c c c c l@{}}  % @{} removes left/right outer padding
\toprule
\textbf{Dataset} & \textbf{Regime} & \textbf{Baseline} & \textbf{Best} &
$\boldsymbol{\Delta}$ & $\boldsymbol{\Delta(\%)}$ &
\textbf{\shortstack{Best gen.\\@ Aug}} \\
\midrule
Cityscapes Pedestrian & From-scratch & 0.4562 & 0.4910 & +0.0348 & +7.6\%  & DiffusionGAN@75\% \\
COCO PottedPlant      & From-scratch & 0.1599 & 0.2088 & +0.0489 & +30.6\% & DiffusionGAN@100\% \\
Traffic Signs         & From-scratch & 0.9297 & 0.9434 & +0.0137 & +1.5\%  & DiT@25\% \\
\midrule
Cityscapes Pedestrian & Pretrained   & 0.6430 & 0.6516 & +0.0086 & +1.3\%  & ADM@25\% \\
COCO PottedPlant      & Pretrained   & 0.4919 & 0.4980 & +0.0061 & +1.2\%  & DiffusionGAN@10\% \\
Traffic Signs         & Pretrained   & 0.9949 & 0.9949 & +0.0000 & +0.0\%  & StyleGAN2-ADA@25\% \\
\bottomrule
\end{tabular}%
}
\endgroup

\end{table}
\noindent\textbf{Cityscapes Pedestrian (dense, occlusion-heavy; held-out test).} On the held-out test split, the real-only baseline is mAP $=0.4562$. Synthetic augmentation improves performance in this regime, with the best observed configuration reaching mAP $=0.4910$ (DiffusionGAN at 75\% augmentation), corresponding to an absolute gain of $\Delta=+0.0348$ ($+7.6\%$ relative). Across generators, mAP typically improves from low to moderate augmentation ratios and then shows diminishing returns or generator-dependent fluctuations at higher ratios. 

\noindent\textbf{COCO PottedPlant (multi-instance, wide scale variation, diverse context; held-out test).} On the held-out test split, the real-only baseline is mAP $=0.1599$. Synthetic augmentation yields the largest absolute gain among the three datasets, with the best observed configuration reaching mAP $=0.2088$ (DiffusionGAN at 100\%), i.e., $\Delta=+0.0489$ ($+30.6\%$ relative). The mean trend across generators increases with augmentation over a wider range than in the other datasets, suggesting that additional diversity is particularly beneficial in this regime.

\noindent\textbf{Traffic Signs (sparse, near-saturated; held-out test).} On the held-out test split, the real-only baseline is mAP $=0.9297$. Augmentation yields comparatively modest changes: the best observed configuration reaches mAP $=0.9434$ (DiT at 25\%), i.e., $\Delta=+0.0137$ ($+1.5\%$ relative). At very high synthetic fractions, some generators slightly underperform the baseline, consistent with saturation and mild distribution-shift effects.

\reshead{Pretrained fine-tuning behavior (held-out test evaluation)}
\label{sec:results-pretrained-summary}

We also evaluate YOLOv11 under COCO-pretrained initialization. Relative to training from scratch, gains (when present) tend to occur at low-to-moderate augmentation ratios, while very large synthetic fractions can be neutral or detrimental depending on dataset and generator. Complete pretrained \emph{training-time validation} curves are reported in Appendix~\ref{app:pretrained_curves}, and full pretrained \emph{held-out test} augmentation curves are reported in Appendix~\ref{app:test_aug_curves}.

\noindent\textbf{Cityscapes Pedestrian (pretrained; held-out test).} The pretrained baseline is mAP $=0.6430$ and the best augmented configuration reaches mAP $=0.6516$ (ADM at 25\%; $\Delta=+0.0086$, $+1.3\%$).

\noindent\textbf{COCO PottedPlant (pretrained; held-out test).} The pretrained baseline is mAP $=0.4919$ and the best augmented configuration reaches mAP $=0.4980$ (DiffusionGAN at 10\%; $\Delta=+0.0061$, $+1.2\%$), though our targeted multi-seed check indicates that effects of this magnitude can fall within run-to-run variability (Appendix~\ref{app:seed_robustness}).

\noindent\textbf{Traffic Signs (pretrained; held-out test).} The pretrained baseline is mAP $=0.9949$ and augmentation produces negligible changes across ratios and generators (best mAP $=0.9949$ with StyleGAN2-ADA at 25\%; $\Delta<10^{-4}$).

\noindent\textbf{Seed robustness.} Because YOLO training has stochastic components, we reran three representative baseline--augmented comparisons with seeds $\{0,1,2\}$ under the identical training recipe and evaluated held-out real test mAP@0.50:0.95. We report $\Delta$ as (Augmented$-$Baseline) and summarize variability as mean$\pm$std over seeds with a Student-$t$ 95\% CI (df$=2$). In the pretrained regime (Pedestrian: baseline vs.\ ADM@25\%; PottedPlant: baseline vs.\ DiffusionGAN@10\%), the 95\% CIs include $0$, indicating that effects of this magnitude are not reliably separable from seed variation. In contrast, the from scratch Traffic Signs comparison remains consistently positive ($\Delta=+0.0131\pm0.0009$, 95\% CI $[0.0108,\,0.0153]$). Full details are in Appendix~\ref{app:seed_robustness}.

\subsection{Metric--performance alignment}
\label{sec:results-metric-alignment}

We next test whether synthetic-dataset metrics computed for each dataset--generator--augmentation configuration under our matched-size bootstrap protocol (Step~2) are informative about downstream YOLOv11 performance (Step~3), using the correlation analyses in Step~4. We group metrics into three families: (i) Inception-based global metrics (e.g., FID, precision/recall, density/coverage, authenticity), (ii) DINOv2-based global metrics (e.g., FD/KD/FD$_\infty$, authenticity and CT-style scores), and (iii) object-centric distribution metrics computed from bounding-box statistics using Wasserstein/Jensen--Shannon distances. For an intuitive point-level view of selected object-centric metrics versus held-out test mAP (including potential non-linearities and outliers), see Appendix~\ref{app:scatter_plots}.

Raw correlation heatmaps are in Appendix~\ref{app:full_heatmaps}. Because raw correlations mix the augmentation--performance trend with generator-to-generator differences at fixed augmentation, we emphasize augmentation-controlled (residualized) correlations in the main text.

\textbf{Decision-oriented screening at fixed budgets.} To complement correlation analyses, we report a decision-style fixed-budget generator screening evaluation (Top-1 accuracy, regret, Kendall $\tau$ at 25/50/100\% budgets) in Appendix~\ref{app:decision_screening} (Table~\ref{tab:decision_screening_summary}); screening signal is clearest in the from scratch PottedPlant regime, while under pretrained fine-tuning generator separation is smaller and screening is correspondingly weaker in this benchmark.

\reshead{Why residual correlation analysis is necessary}
\label{sec:results-residual-motivation}

A key confound is that the \emph{augmentation ratio} can affect both dataset metrics (because the synthetic pool changes with augmentation level) and mAP (because the effective training set size and distribution change). Therefore, naive metric--mAP correlations can reflect shared dependence on augmentation ratio rather than a metric's ability to distinguish which \emph{generator} is better at a fixed augmentation budget.

To evaluate association \emph{beyond augmentation amount}, we report: (i) \emph{raw} metric--mAP correlations, and (ii) \emph{residual} correlations after regressing out augmentation ratio. Concretely, for each dataset and regime, we fit $mAP_{\text{test}}(a)=\alpha+\beta a+\epsilon$ and, for each metric $M$, fit $M(a)=\gamma+\delta a+\eta$, where $a$ is augmentation ratio. We then compute correlation between the residuals $(\epsilon,\eta)$, which tests whether $M$ explains variation in held-out test mAP beyond the augmentation-level trend. Because we test many metrics, we correct for multiple comparisons using the Benjamini--Hochberg false discovery rate (BH--FDR) procedure~\citep{Benjamini1995}. We report significance using $q$-values (FDR-controlled), and mark cells with $q<0.05$. Comparing the raw heatmaps (Appendix~\ref{app:full_heatmaps}) to the residual heatmaps (Figure~\ref{fig:heatmap_resid_fromscratch_key} and Appendix~\ref{app:full_heatmaps} and Appendix~\ref{app:pretrained_heatmaps}) shows that augmentation control can materially change which metric--dataset pairs appear associated with mAP and which associations survive BH--FDR correction. Because augmentation–performance curves can be non-linear and each generator is evaluated across multiple augmentation levels, we additionally validate augmentation control using per-level correlations and a categorical augmentation fixed-effects model. Results are consistent with the main residual analysis (Appendix~\ref{app:residualization_robustness}).

\reshead{Residual correlation heatmaps (From-Scratch)}
\label{sec:results-residual-heatmaps}

Figure~\ref{fig:heatmap_resid_fromscratch_key} summarizes residual Spearman correlations (augmentation trend removed) for a compact set of key metrics spanning Inception-based, DINOv2-based, and object-centric families. For example, in COCO PottedPlant, lower \texttt{kd\_value\_inception} and \texttt{sw\_approx\_inception} align with higher mAP ($\rho=-0.47$ and $-0.45$), while Traffic Signs exhibits weaker residual structure. For readability, we visualize a subset of K=15 `key metrics', selected by ranking metrics according to the \emph{largest} absolute residual Spearman correlation they attain in any of the three datasets (i.e., the strongest-magnitude cell per metric in the residual correlation matrix) and taking the top-K unique metrics. This effect-size-based selection is used only to keep Figure~\ref{fig:heatmap_resid_fromscratch_key} compact. Statistical significance is indicated \emph{separately} via BH--FDR-corrected $q$-values: we compute BH--FDR on the full set of correlation tests and mark cells with $q<0.05$ using asterisks. A full residual-correlation heatmap over the complete metric suite is provided in Appendix~\ref{app:full_heatmaps}. 
{\setlength{\textfloatsep}{8pt}
\begin{figure}[t]
    \centering
    \includegraphics[width=0.97\linewidth]{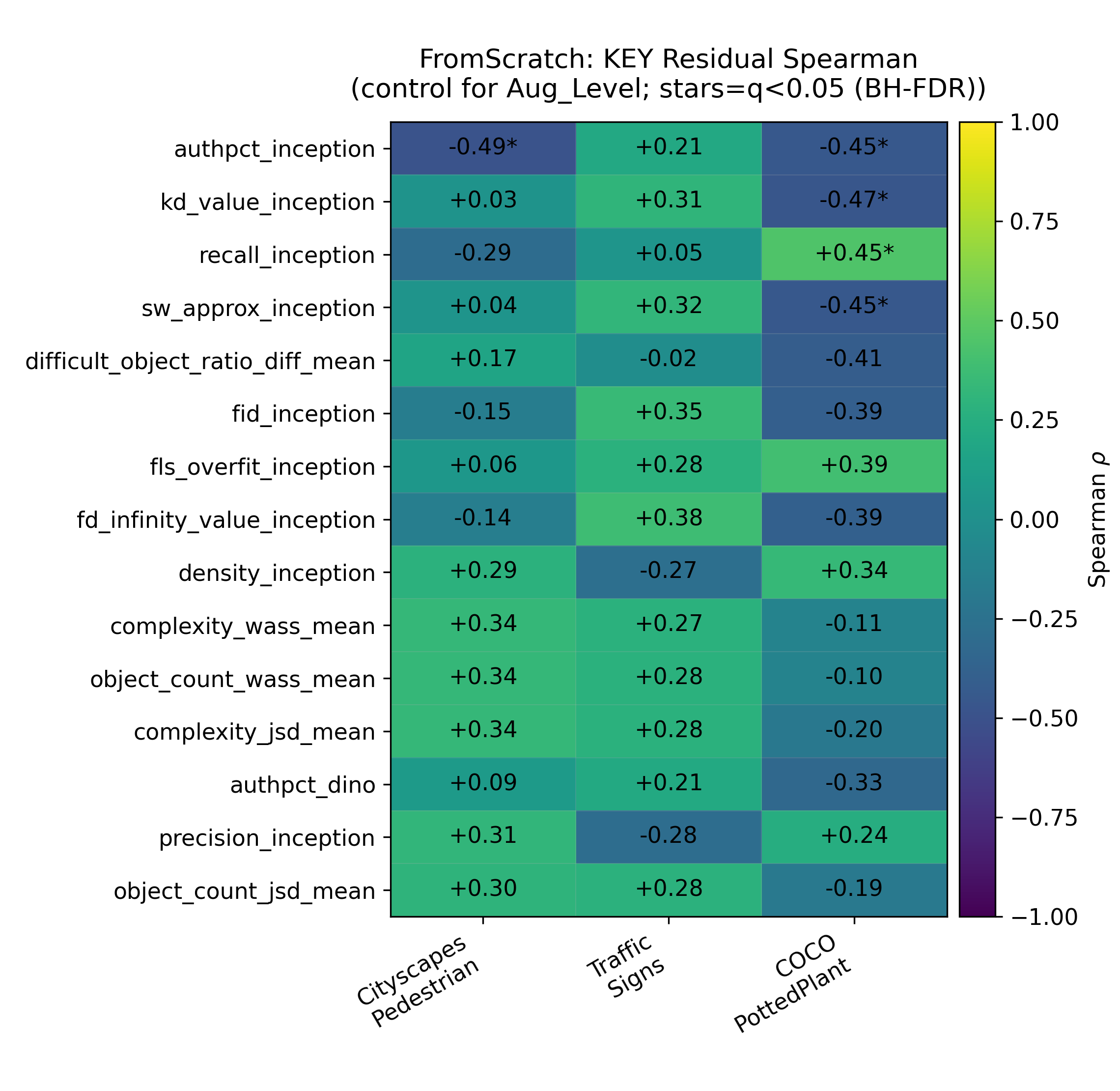}
    \caption{\small Residual Spearman correlations between synthetic-data metrics and YOLOv11 mAP@0.50:0.95 for the From-Scratch regime, controlling for augmentation ratio. Asterisks denote BH--FDR corrected significance at $q<0.05$ \citep{Benjamini1995}.}
    \label{fig:heatmap_resid_fromscratch_key}
\end{figure}
}
\paragraph{From-Scratch: metric signal is dataset-dependent.}
Only a subset of metrics exhibit statistically significant residual associations after BH--FDR correction, and the sign/magnitude of correlations varies across datasets. This indicates that metric utility is \emph{not universal}: some global feature-space metrics (Inception/DINOv2) track generator-dependent performance differences beyond the augmentation trend in certain datasets, while object-centric distribution metrics appear most informative in regimes where performance is sensitive to object difficulty statistics.

\reshead{Pretrained residual correlations}
\label{sec:results-residual-pretrained}
We repeat the residual-correlation analysis for COCO-pretrained fine-tuning. Because pretrained detectors can be less sensitive to some forms of synthetic mismatch (while still being affected by larger distribution shifts), the residual correlation structure differs from the from scratch case. Pretrained residual heatmaps are reported in Appendix~\ref{app:pretrained_heatmaps}; in this regime, residual correlations are generally weaker and rarely survive BH--FDR correction (with only an isolated significant association visible in our full-suite residual heatmap).

These results show that the augmentation effects and metric--performance alignment are strongly regime-dependent, motivating the discussion in Section~\ref{sec:discussion_conclusion}. We discuss the main empirical patterns from Section~\ref{sec:results}, clarify what the correlation analyses do (and do not) support, and close the loop on Q1--Q3.

\section{Discussion and Conclusions}
\label{sec:discussion_conclusion}
\noindent\textbf{Synthetic augmentation effects are regime- and initialization-dependent.} Across the three datasets, synthetic augmentation can improve YOLOv11 mAP, but the magnitude and consistency of improvements depend on the detection regime and initialization (Figure~\ref{fig:aug_curves_fromscratch}, Table~\ref{tab:best_map_summary}). In the from scratch setting, the largest gains appear in regimes with clear headroom (dense/occlusion-heavy scenes and high-variability multi-instance scenes), whereas the near-saturated Traffic Signs regime exhibits smaller changes and occasional degradation at high synthetic fractions. Under COCO-pretrained fine-tuning, improvements (when present) are  concentrated at low-to-moderate augmentation ratios, and generator-to-generator differences are smaller, consistent with pretraining providing a strong representation prior while making distribution shift more consequential when synthetic images dominate the training mix (Appendix~\ref{app:pretrained_curves}).

\noindent \textbf{Metric-performance alignment beyond augmentation: residual analysis with multiple-testing control.}
A key confound is that augmentation ratio can influence both metric values and mAP, so raw metric--mAP correlations may largely reflect the shared dependence on augmentation level rather than a metric’s ability to rank generators at a fixed synthetic-data budget. To target the screening use-case, we therefore report both raw correlations and augmentation-controlled (residualized) correlations (Section~\ref{sec:results-residual-motivation}, Appendix~\ref{app:residual_correlation_details}). Residualization changes the picture: some raw associations weaken, and only some metric--dataset pairs remains detectable. Because many metric--dataset correlations are tested in parallel, we control the false discovery rate using BH--FDR on the full set of correlation tests within each regime and correlation view (raw vs.\ residual), and report significance via $q$-values (Appendix~\ref{app:bh_fdr_details})~\citep{Benjamini1995}. Overall, the pattern is not universal: which metrics align with downstream mAP (and in what direction) varies across regimes and initialization, and metric-based generator prioritization should be treated as dataset-specific rather than a one-size-fits-all rule (Figure~\ref{fig:heatmap_resid_fromscratch_key}).

\noindent \textbf{Limitations and scope.}
Our analysis is associative rather than causal. Residualizing by augmentation removes the dominant budget trend, but other confounders (e.g., generator artifacts, annotation noise, dataset bias) may remain, and a linear adjustment may miss non-linear augmentation effects. Because each generator appears at multiple augmentation levels, data points are not independent; significance should be interpreted as evidence within this experimental grid rather than guaranteed transfer. Results are specific to YOLOv11 and three single-class datasets; future work should test multi-class and other detector families.

\noindent \textbf{Answering the study questions (Q1--Q3) and implications for practice.}
\textbf{(Q1)} Global, encoder-based metrics (Inception-v3 and DINOv2) provide regime-specific screening signal within this benchmark, but no single metric is consistently predictive across all three datasets and both initialization regimes once augmentation ratio is controlled (Figure~\ref{fig:heatmap_resid_fromscratch_key}, Appendix~\ref{app:full_heatmaps}, Appendix~\ref{app:pretrained_heatmaps}).
\textbf{(Q2)} Object-centric distribution metrics over bounding-box statistics provide complementary, interpretable signal about label-space mismatch (e.g., instance density and difficult/small-object prevalence) that global feature-space metrics do not explicitly target. In our from scratch results, \texttt{difficult\_object\_ratio\_diff\_mean} (gap in \emph{small/occluded-object prevalence}; Appendix~\ref{app:metric_definitions}) shows a detectable fixed-effects association in Pedestrian and PottedPlant (Table~\ref{tab:residualization_robustness}), but remains regime-dependent after controlling for augmentation; we therefore treat object-centric metrics primarily as diagnostics rather than a universal ranking rule.
\textbf{(Q3)} Metric--performance relationships vary with both dataset regime and training regime (from scratch vs.\ pretrained). Practically, augmentation is most likely to help when the baseline leaves headroom; near-saturated regimes show limited upside and can be sensitive to distribution shift at high synthetic fractions. For generator choice at a fixed augmentation budget, residual (augmentation-controlled) analysis directly targets generator-to-generator mAP differences at the same budget, and the decision-style screening results are strongest in the from scratch PottedPlant setting but weaker under pretrained fine-tuning in this benchmark (Appendix~\ref{app:decision_screening}).

\noindent \textbf{Conclusions.}
Synthetic augmentation can improve YOLOv11 performance, but the magnitude and stability of gains depend strongly on detection regime and initialization, with the largest headroom typically in from scratch settings where a suitable pretrained model may be unavailable or unreliable. With respect to \emph{pre-training} metric-based screening, augmentation-controlled (residualized) correlation analysis with BH--FDR correction suggests that metric--performance alignment is not universal: only a subset of metrics shows detectable association with downstream mAP beyond the dominant effect of augmentation ratio, and which metrics align varies across regimes. To target the screening use-case (choosing a generator and augmentation ratio \emph{before} YOLO training), we report both raw correlations and residualized correlations: raw trends capture how metrics co-vary with performance as synthetic volume increases, while residualization tests whether metrics help distinguish \emph{which generator} performs better at a fixed synthetic-data budget. Across our three regimes, the residual heatmaps  provide \emph{dataset-specific, exploratory guidance} for prioritizing generators/ratios, but we caution against treating any single global metric as a universal predictor across datasets.

\section*{Impact Statement}
This paper studies synthetic data augmentation for object detection and evaluates whether
pre-training dataset metrics predict downstream YOLO performance. The work can reduce
compute and data costs by helping practitioners choose augmentation strategies more
efficiently, but it may also enable faster deployment of detectors in sensitive settings.
We do not introduce new surveillance capabilities, but improved detector training pipelines
could be misused. We encourage responsible use and evaluation on domain-appropriate data,
and reporting of failure modes and bias where applicable.

\section*{Reproducibility Statement}
We will release a public repository containing code and artifacts to reproduce all figures
and analyses in this paper, including processed results, split definitions, and (where
licensing permits) the labeled synthetic datasets. We will update the arXiv record with
the repository URL when it becomes available.

\bibliography{main}
\bibliographystyle{bibstyl}

%%%%%%%%%%%%%%%%%%%%%%%%%%%%%%%%%%%%%%%%%%%%%%%%%%%%%%%%%%%%%%%%%%%%%%%%%%%%%%%
%%%%%%%%%%%%%%%%%%%%%%%%%%%%%%%%%%%%%%%%%%%%%%%%%%%%%%%%%%%%%%%%%%%%%%%%%%%%%%%
% APPENDIX
%%%%%%%%%%%%%%%%%%%%%%%%%%%%%%%%%%%%%%%%%%%%%%%%%%%%%%%%%%%%%%%%%%%%%%%%%%%%%%%
%%%%%%%%%%%%%%%%%%%%%%%%%%%%%%%%%%%%%%%%%%%%%%%%%%%%%%%%%%%%%%%%%%%%%%%%%%%%%%%
%\newpage
%\appendix
%\onecolumn

\clearpage
\appendix
\setcounter{section}{0}
%\maketitlesupplementary

% ------------------------------------------------------------
% Supplementary Sec. A  (referenced in the main paper)
% ------------------------------------------------------------
\section{Supplementary dataset examples and regime statistics}
\label{sec:supp-dataset-samples}
 \subsection{Dataset curation and preprocessing details}
\label{app:dataset_curation_details}

\noindent\textbf{Cityscapes Pedestrian.}
This dataset is derived from Cityscapes~\citep{Cordts2016Cityscapes}. We select images containing pedestrian-related categories and merge them into a single class: person, rider, and sitting person are mapped to \emph{pedestrian}. The original frames have resolution $2048\times1024$. To emphasize dense arrangements, we crop each frame into $768\times768$ patches centered on regions with high pedestrian density and occlusion. The final curated set contains 2195 images.

\noindent\textbf{Traffic Signs.}
This dataset represents a simpler regime with high-contrast signage and typically low overlap. The source data contains 4185 images with 4883 labeled instances (original resolution $1024\times768$). For consistency with the other domains and to control real-data volume, we curate a 2568-image subset. Each selected image is cropped to $768\times768$ and all sign-related categories are merged into a single \emph{traffic sign} class.

\noindent\textbf{COCO PottedPlant.}
We construct a single-class subset from MS-COCO~\citep{Lin2014COCO} using the \emph{potted plant} category. The extracted subset contains 4624 images with varying resolutions (commonly around $640\times480$). We standardize preprocessing by resizing and cropping to a fixed $512\times512$ resolution, and we curate a final set of 2568 images.

\subsection{YOLOv11 training details}
\label{app:yolo_training_details}

Across all datasets, we use the same YOLOv11 training recipe within a given initialization regime
(from scratch or COCO-pretrained). Unless stated otherwise, training uses cosine learning-rate scheduling
with initial learning rate 0.001, batch size 32, momentum 0.67, weight decay 0.001, and early stopping
patience 50 epochs. We train for up to 600 epochs and select the checkpoint with the best validation mAP
for held-out test evaluation. Image sizes are \texttt{imgsz}=768 for Cityscapes Pedestrian and Traffic Signs,
and \texttt{imgsz}=512 for COCO PottedPlant.

\subsection{Additional real dataset examples}
\label{app:dataset_examples}
This section provides additional real images from each dataset after preprocessing, complementing Figure~\ref{fig:SamplesGrid} in the main paper. The examples illustrate the three detection regimes studied: (i) sparse, near single-object scenes (Traffic Signs), (ii) dense scenes with frequent occlusion and many small objects (Cityscapes Pedestrian), and (iii) multi-instance scenes with wide object-scale variation (often many small instances and occasional overlaps) and highly diverse indoor/outdoor backgrounds (COCO PottedPlant). All examples show the \emph{real} training-domain images after the same preprocessing used for YOLO training. 

\begin{figure}[ht]
    \centering
    \includegraphics[width=0.90\linewidth]{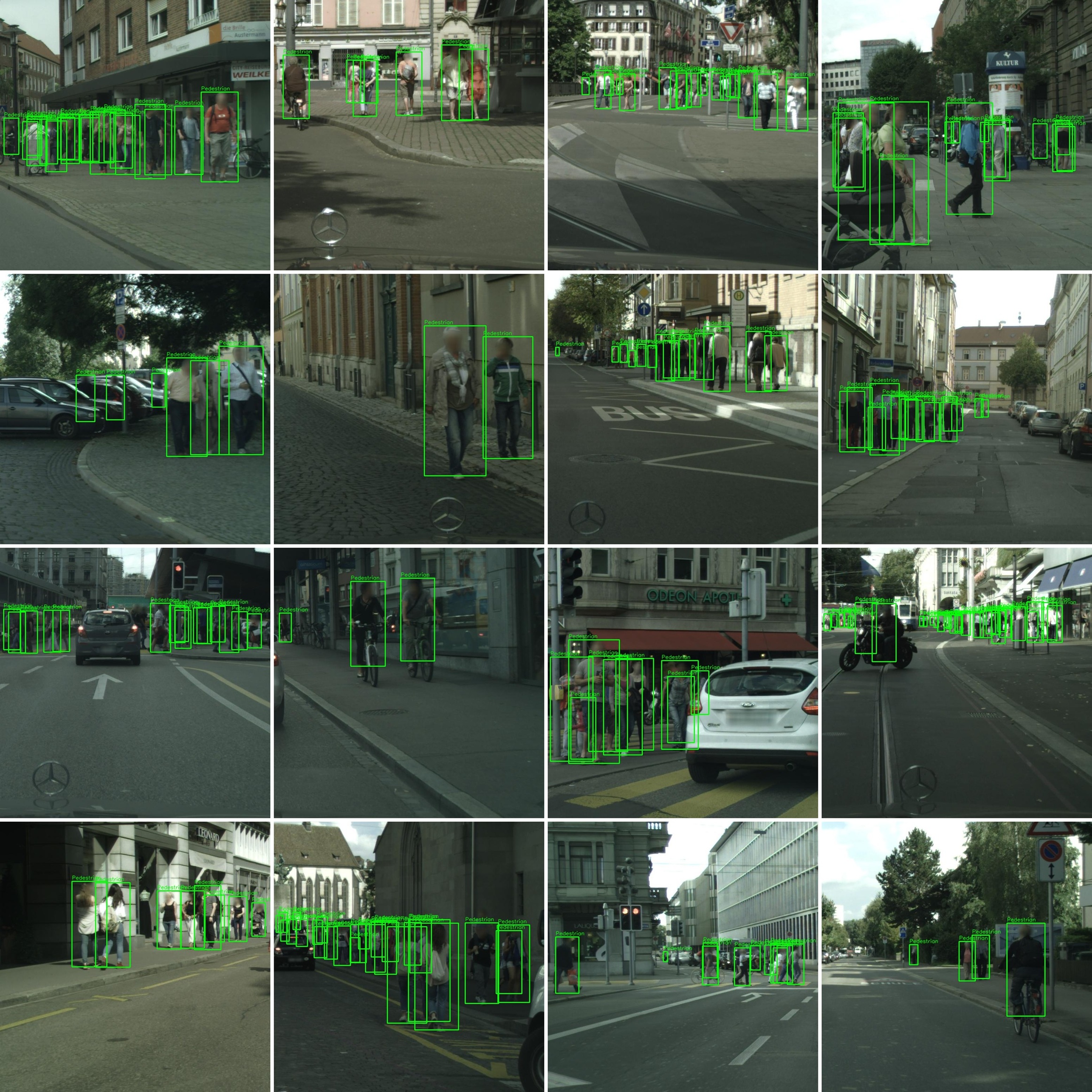}
    \caption{\small Additional real samples from \textbf{Cityscapes Pedestrian}. Scenes are dense with frequent occlusion and multiple small objects.}
    \label{fig:PedestrianSup}
\end{figure}

\begin{figure}[ht]
    \centering
    \includegraphics[width=0.90\linewidth]{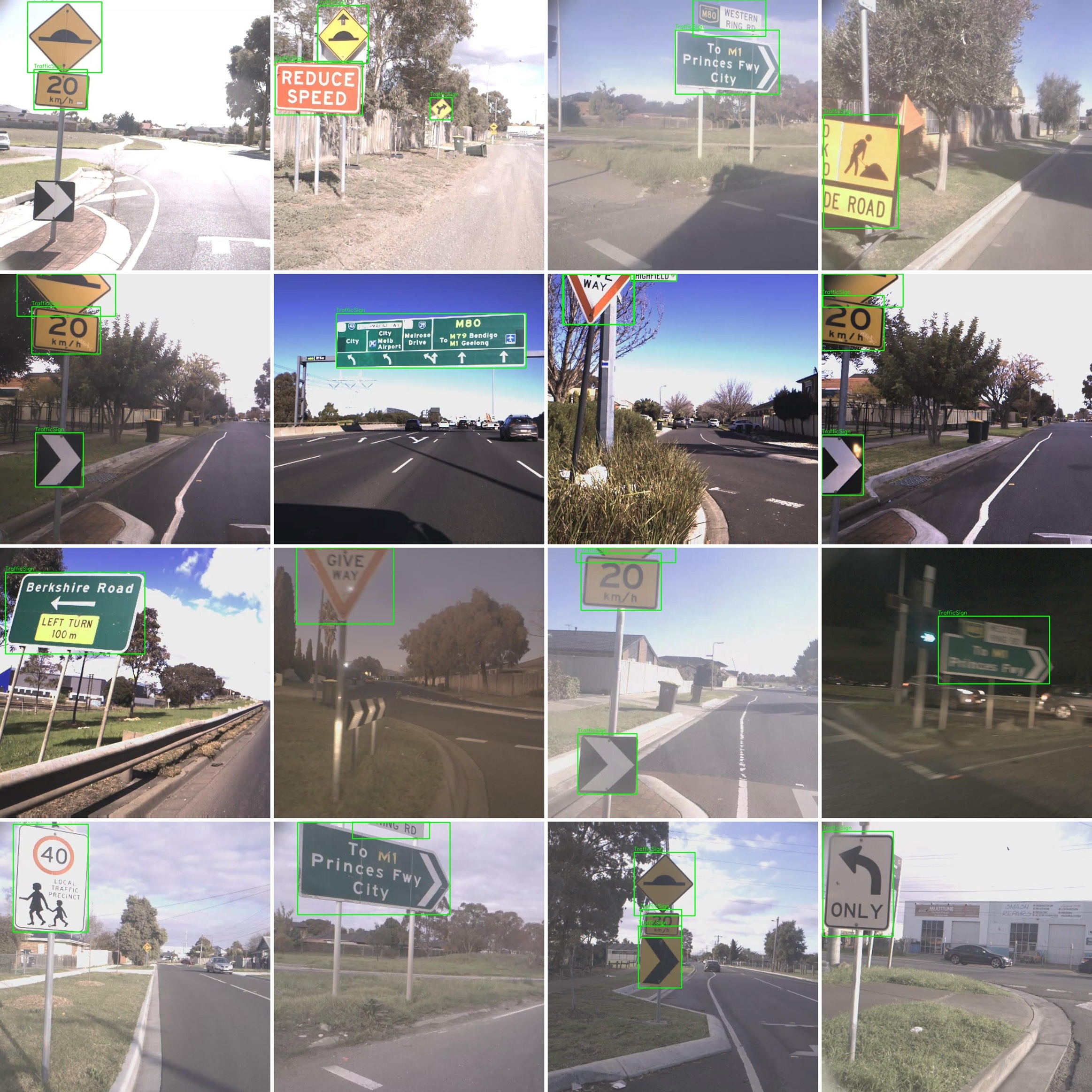}
    \caption{\small Additional real samples from \textbf{Traffic Signs}. Most images contain a single dominant object with minimal overlap.}
    \label{fig:TrafficSignSup}
\end{figure}

\begin{figure}[ht]
    \centering
     \includegraphics[width=0.90\linewidth]{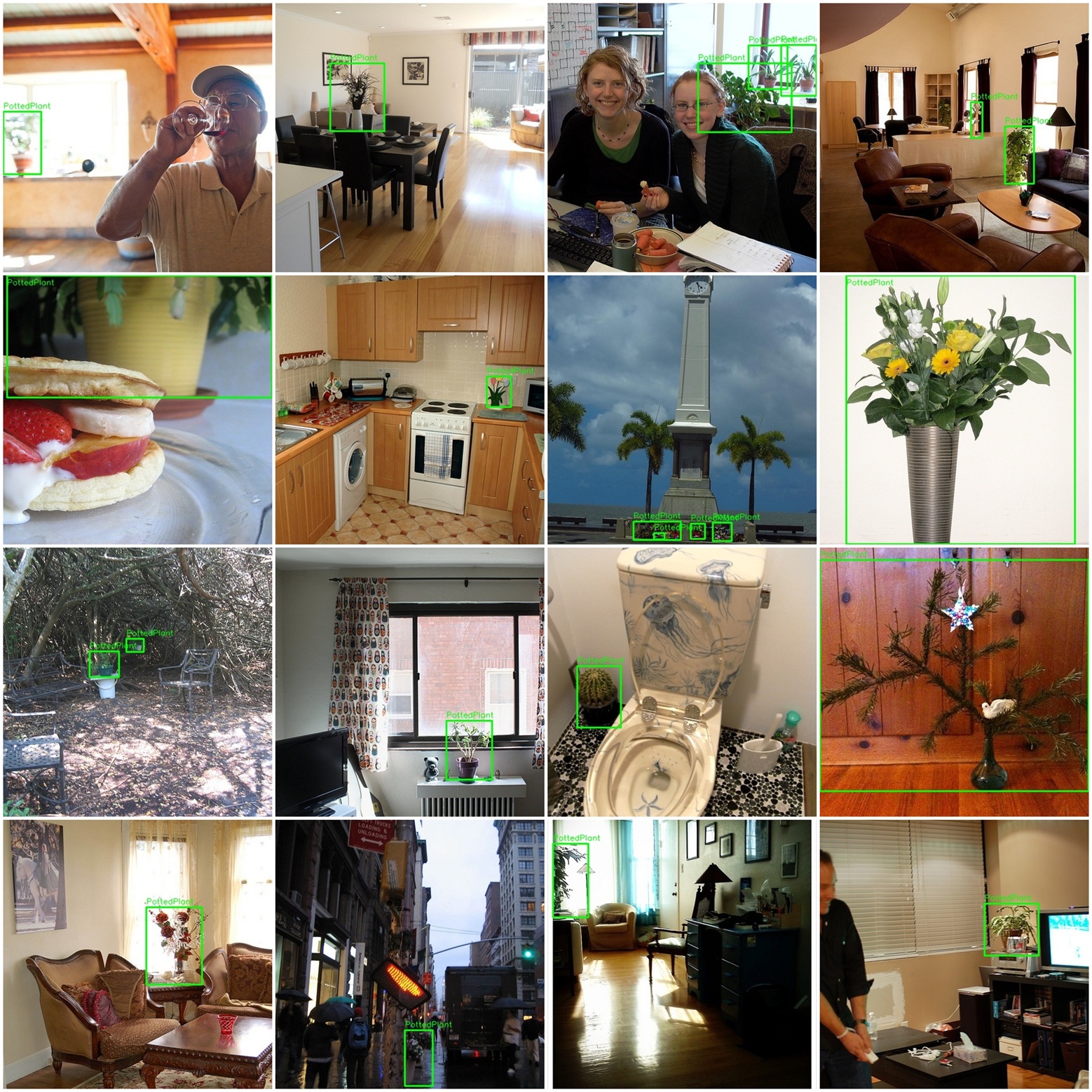}
    \caption{\small Additional real samples from COCO PottedPlant. Images often contain multiple instances with wide scale variation (including many small objects) and occasional overlaps, with highly diverse indoor/outdoor backgrounds.}
    \label{fig:PottedPlantSup}
\end{figure}

\subsection{Dataset regime statistics (real training splits)}
\label{app:dataset_regime_stats}

To support the regime interpretation used throughout the paper, we summarize label-derived statistics computed on the \emph{real training splits after preprocessing}. We report (i) the distribution of instance counts per image, (ii) the prevalence of small objects, (iii) the typical box scale (normalized box area), and (iv) the degree of within-image overlap (pairwise IoU). These quantities directly reflect the object-centric properties targeted by the Wasserstein/JSD metrics in Section~\ref{sec:results-metric-alignment}.
\begin{table}[t]
\centering
\caption{\small Regime statistics on real training splits after preprocessing. ``Small'' denotes boxes with normalized area $<0.01$ (i.e., $<1\%$ of image area). Mean pairwise IoU is computed per image over all box pairs (images with fewer than two boxes contribute IoU $=0$), then averaged over images.}
\label{tab:dataset_regime_stats}

\footnotesize
\setlength{\tabcolsep}{3.5pt}
\renewcommand{\arraystretch}{1.05}

\resizebox{\columnwidth}{!}{%
\begin{tabular}{@{}lcccc@{}}
\toprule
\textbf{Dataset} & \textbf{Inst./img} & \textbf{\% small} & \textbf{Mean area} & \textbf{Mean IoU} \\
\midrule
Cityscapes Pedestrian & $5.08\pm2.80$ & 67.80\% & $0.014\pm0.008$ & $0.041\pm0.054$ \\
Traffic Signs         & $2.95\pm1.62$ & 8.91\%  & $0.032\pm0.023$ & $0.008\pm0.023$ \\
COCO PottedPlant      & $1.80\pm1.14$ & 36.79\% & $0.113\pm0.168$ & $0.009\pm0.029$ \\
\bottomrule
\end{tabular}%
}
\end{table}

\paragraph{Interpretation (regime differences).}
Cityscapes Pedestrian exhibits the highest instance density ($5.08\pm2.80$ instances/img) and the strongest small-object dominance (67.8\% of boxes occupy $<1\%$ of image area), together with the highest overlap proxy (mean pairwise IoU $=0.041$). This is consistent with its role as the most occlusion-heavy regime. Traffic Signs is comparatively sparse ($2.95\pm1.62$ instances/img), with a low fraction of small boxes (8.9\%) and negligible overlap (IoU $=0.008$), matching the ``sparse, low-overlap'' regime used in the main text. COCO PottedPlant shows \emph{multi-instance} structure with substantial variability ($1.80\pm1.14$ instances/img) and a pronounced small-object component (36.8\% small boxes), while also exhibiting the widest scale variation (mean box area ratio $0.113$ with high dispersion), reflecting that the object class appears across a wide range of apparent sizes and diverse indoor/outdoor contexts. Overall, these statistics support the paper’s regime narrative: Pedestrian is overlap- and small-object dominated, Traffic Signs is sparse and clean, and PottedPlant combines diverse context with frequent small instances and broad scale variability, which helps explain why augmentation behavior and metric--performance alignment differ across the three datasets.

\section{Additional results and diagnostics}
\label{sec:supp-training-curves}

\subsection{Pretrained: full per-generator curves} 
\label{app:pretrained_curves}

\begin{figure*}[t]
    \centering
    \includegraphics[width=0.98\textwidth]{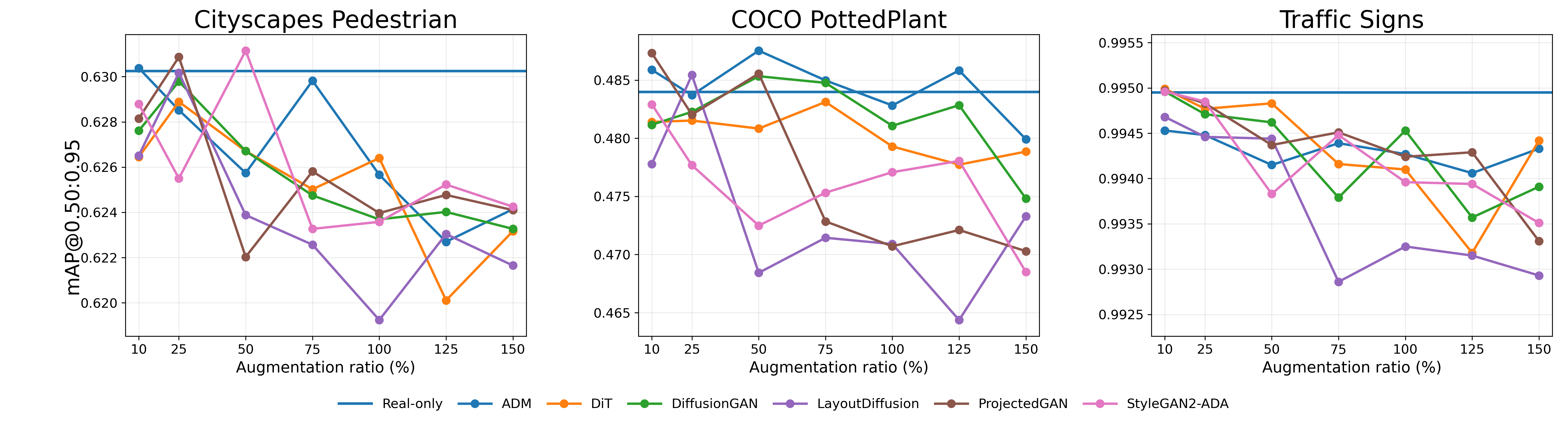}
    \caption{\small YOLOv11 \textbf{Pretrained} \emph{training-time validation} mAP@0.50:0.95 vs.\ augmentation ratio (per-generator curves). Curves report the \emph{best validation mAP observed during training} (from Ultralytics training logs); the horizontal line indicates the real-only baseline.}
    \label{fig:aug_curves_pretrained_full}
\end{figure*}

Figure~\ref{fig:aug_curves_pretrained_full} provides the complete per-generator augmentation curves for COCO-pretrained fine-tuning,
as referenced in Section~\ref{sec:results-pretrained-summary} of the main paper.

\subsection{Held-out test evaluation augmentation curves (from scratch and pretrained)}
\label{app:test_aug_curves}

\begin{figure*}[t]
    \centering
    \includegraphics[width=0.98\textwidth]{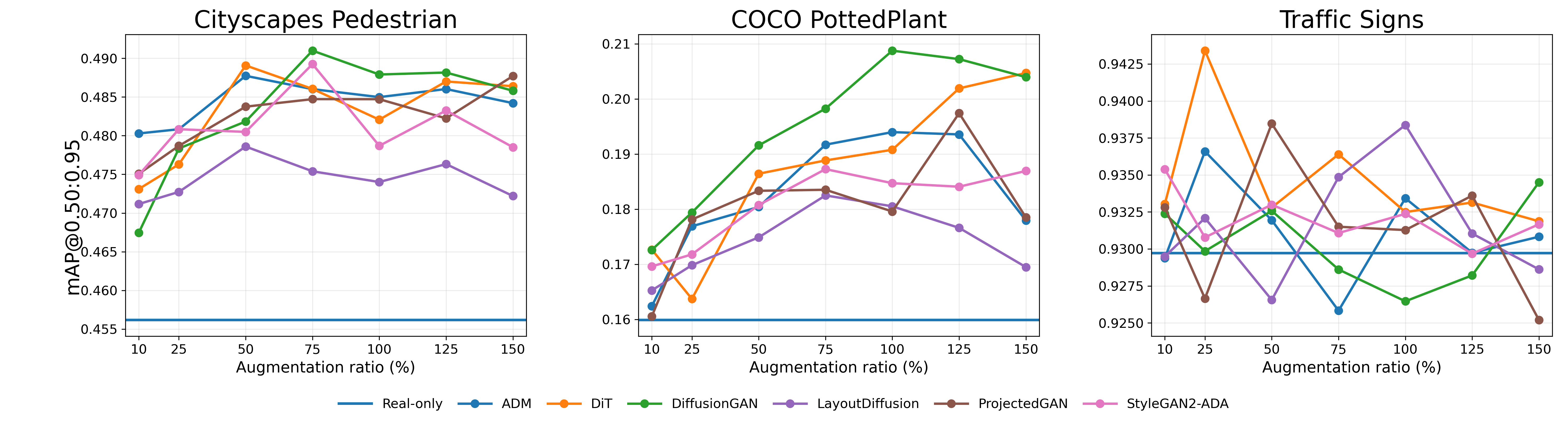}
    \caption{\small YOLOv11 \textbf{From-scratch} \emph{held-out test evaluation} mAP@0.50:0.95 vs.\ augmentation ratio (Ultralytics \texttt{val} on the real test split). Each curve corresponds to one generator; the horizontal line indicates the real-only baseline.}
    \label{fig:aug_curves_test_fromscratch_full}
\end{figure*}

\begin{figure*}[t]
    \centering
    \includegraphics[width=0.98\textwidth]{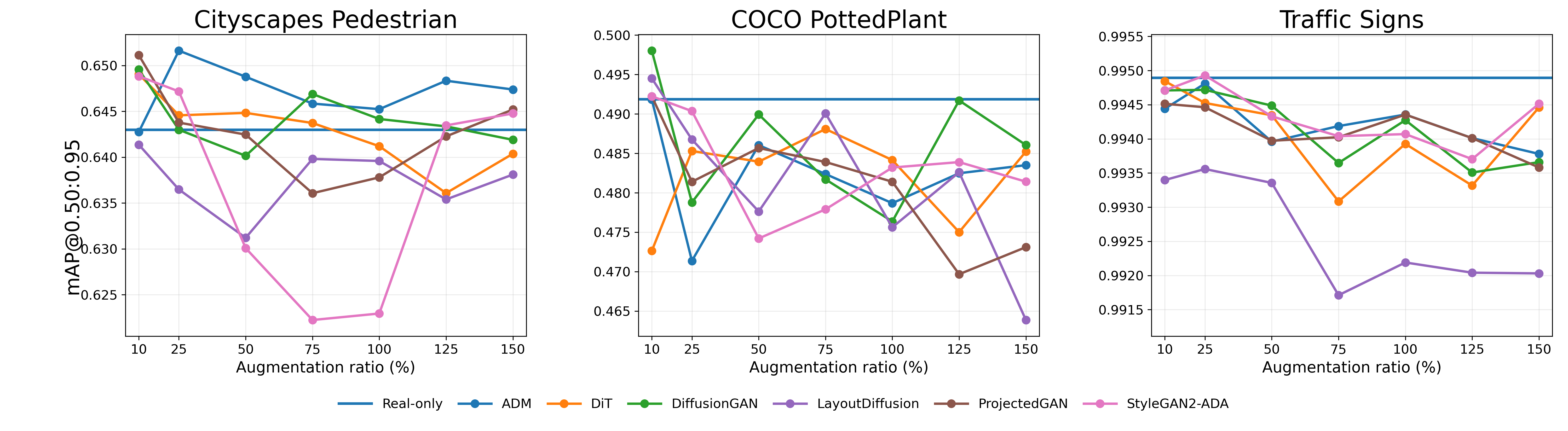}
    \caption{\small YOLOv11 \textbf{Pretrained} \emph{held-out test evaluation} mAP@0.50:0.95 vs.\ augmentation ratio (Ultralytics \texttt{val} on the real test split). Each curve corresponds to one generator; the horizontal line indicates the real-only baseline.}
    \label{fig:aug_curves_test_pretrained_full}
\end{figure*}

Figures~\ref{fig:aug_curves_test_fromscratch_full} and~\ref{fig:aug_curves_test_pretrained_full} report augmentation curves computed from \emph{held-out test evaluation} using the Ultralytics \texttt{val} script on the unseen real test split. Unlike the training-time validation curves (Figures~\ref{fig:aug_curves_fromscratch} and~\ref{fig:aug_curves_pretrained_full}), these results reflect post-training generalization to the test set.

\paragraph{Interpretation.}
Overall, the held-out test curves broadly mirror the training-time validation trends but are slightly noisier, as expected from evaluation on a separate split. 
In the from scratch regime, augmentation benefits are strongest for Cityscapes Pedestrian and COCO PottedPlant, while Traffic Signs remains near-saturated and exhibits small, generator-dependent fluctuations around the baseline.  Under COCO-pretrained fine-tuning, performance is generally less sensitive to augmentation ratio: improvements (when present) concentrate at low-to-moderate synthetic fractions, and some generators show mild degradation at high augmentation ratios, consistent with pretraining reducing headroom and making distribution shift more consequential when synthetic data dominates.

\paragraph{Relation to Table~\ref{tab:best_map_summary}.}
These curves show the full per-generator test-set response across augmentation ratios; Table~\ref{tab:best_map_summary} extracts the best-performing augmented point per dataset/regime and compares it to the real-only baseline.

\subsection{Seed robustness experiments}
\label{app:seed_robustness}

\paragraph{Motivation.}
YOLO training includes stochastic components (e.g., data shuffling and augmentation sampling), so single-run mAP can vary across random seeds, particularly when improvements are small. The full experimental grid in the main paper is run with a fixed seed (\texttt{seed}=0) under deterministic settings to ensure strict comparability across the many dataset--generator--budget configurations. To address concerns about statistical reliability, we additionally perform a targeted multi-seed robustness check on representative comparisons that (i) correspond to small-delta pretrained settings and/or (ii) are used as key examples in the main text.

\paragraph{Protocol.}
We rerun the same training recipe and evaluation pipeline for three seeds (\texttt{seed}\,$\in\{0,1,2\}$) for the following baseline vs.\ augmented pairs:
(i) pretrained Cityscapes Pedestrian baseline vs.\ ADM@25\%;
(ii) pretrained COCO PottedPlant baseline vs.\ DiffusionGAN@10\%;
and (iii) from scratch Traffic Signs baseline vs.\ DiT@25\%.
For each seed, we record held-out evaluation mAP@0.50:0.95 (Ultralytics \texttt{val}; \texttt{metrics/mAP50-95(B)}).
We compute $\Delta$ per seed as (Augmented$-$Baseline), then report mean$\pm$std over seeds.
We also report a 95\% confidence interval (CI) on $\Delta$ using a Student-$t$ interval with df$=2$ (3 seeds). Results are summarized in Table~\ref{tab:seed_robustness}.

\begin{table}[t]
\centering
\caption{\small Seed robustness on held-out mAP@0.50:0.95 (seeds 0/1/2). Values are mean$\pm$std. $\Delta$ is (Aug$-$Base) per-seed, summarized as mean$\pm$std. CI$_{95}$ uses Student-$t$ (df=2).}
\label{tab:seed_robustness}

\small
\setlength{\tabcolsep}{4pt}
\renewcommand{\arraystretch}{1.12}

\begin{tabular}{@{}p{0.18\columnwidth}p{0.1\columnwidth}p{0.28\columnwidth}p{0.34\columnwidth}@{}}
\toprule
\textbf{Data} & \textbf{Init.} & \textbf{Item} & \textbf{Value} \\
\midrule

COCO& Pretr. & Base mAP & $0.4932\!\pm\!0.0019$ \\
          &        & DiffGAN@10\%& $0.4852\!\pm\!0.0133$ \\
          &        & $\Delta$ (Aug$-$Base)& $-0.0080\!\pm\!0.0151$ \\
          &        & CI$_{95}$ on $\Delta$ & $[-0.0454,\,0.0294]$ \\
\addlinespace[3pt]

Cityscapes& Pretr. & Base mAP & $0.6352\!\pm\!0.0107$ \\
               &        & ADM@25\%& $0.6352\!\pm\!0.0085$ \\
               &        & $\Delta$ (Aug$-$Base)& $+0.0000\!\pm\!0.0110$ \\
               &        & CI$_{95}$ on $\Delta$ & $[-0.0274,\,0.0274]$ \\
\addlinespace[3pt]

Traffic& Scratch & Base mAP & $0.9296\!\pm\!0.0007$ \\
             &        & DiT@25\%& $0.9426\!\pm\!0.0007$ \\
             &        & $\Delta$ (Aug$-$Base)& $+0.0131\!\pm\!0.0009$ \\
             &        & CI$_{95}$ on $\Delta$ & $[0.0108,\,0.0153]$ \\

\bottomrule
\end{tabular}
\end{table}

\paragraph{Takeaway.}
The multi-seed check supports a conservative interpretation of small-delta results: the pretrained Pedestrian and PottedPlant comparisons have confidence intervals that include zero, indicating that effects of this magnitude may be within seed variability. In contrast, the from scratch Traffic Signs comparison shows a consistently positive $\Delta$ across seeds (CI excludes zero), suggesting that this moderate gain is robust to random initialization and training stochasticity. Overall, this aligns with the paper's main message that synthetic augmentation benefits are strongest and most reliable in regimes with meaningful headroom, while very small pretrained deltas should be treated as potentially noisy.

\subsection{Residualized scatter plots (diagnostic; from-scratch)}
\label{app:scatter_plots}

\begin{figure*}[t]
    \centering
    \includegraphics[width=0.98\textwidth]{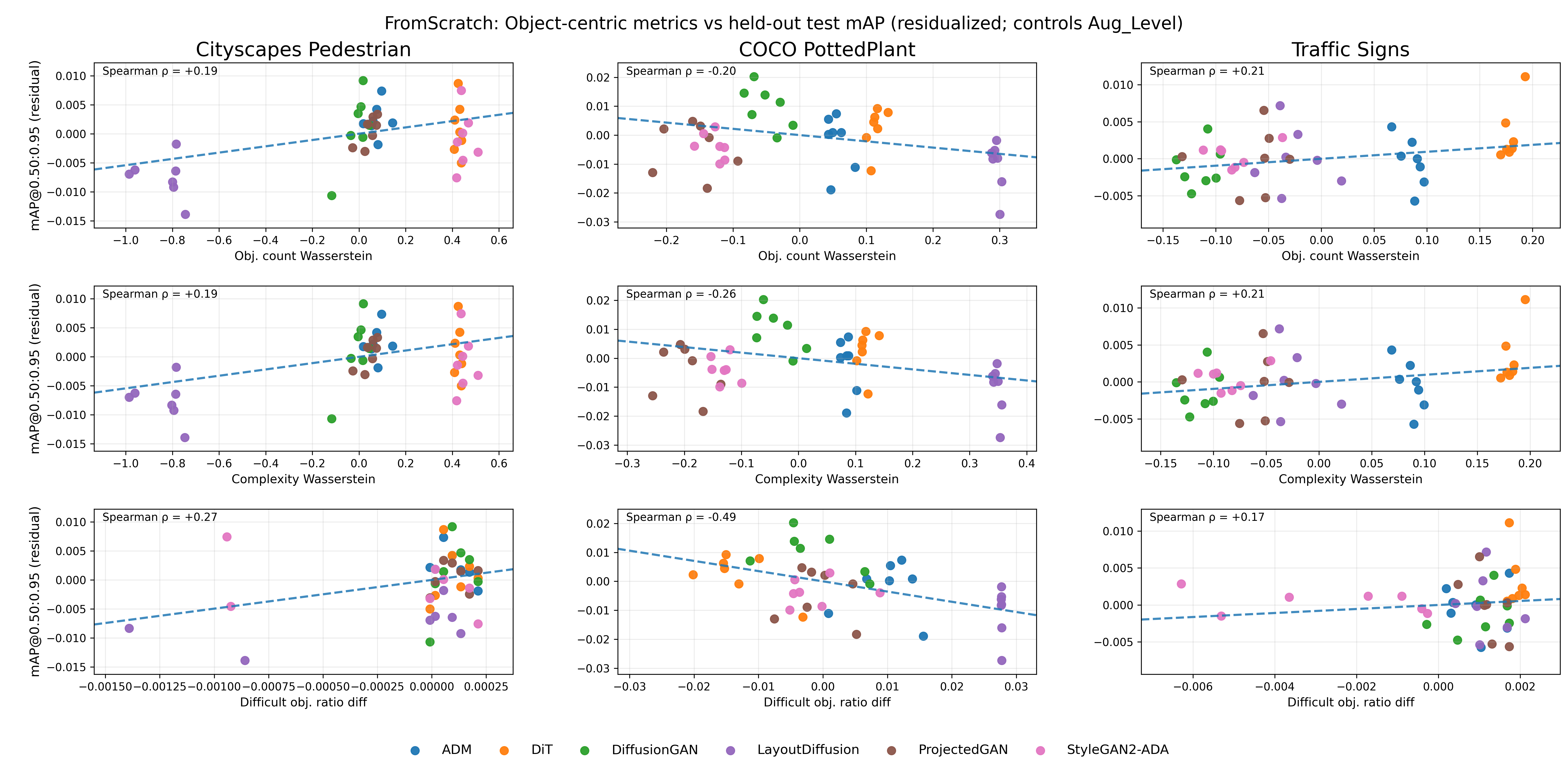}
    \caption{\small Residualized scatter plots (\textbf{From-Scratch}): selected object-centric distribution metrics vs.\ held-out test mAP residuals (augmentation-ratio trend removed). Rows correspond to object-count Wasserstein, complexity Wasserstein, and difficult-object-ratio difference (\emph{raw} $=|r_{\text{real}}-r_{\text{syn}}|\ge 0$); columns correspond to datasets. Dashed line is an OLS fit in residual space; Spearman $\rho$ is reported per panel. These plots are diagnostic: they visualize outliers and generator-specific clustering that a single correlation coefficient can obscure.}
    \label{fig:scatter_objectcentric_fromscratch}
\end{figure*}

Residual correlation heatmaps and fixed-budget screening tables provide the main, decision-relevant summaries of metric--performance alignment. The scatter plot composite in Figure~\ref{fig:scatter_objectcentric_fromscratch} is included only as a diagnostic visualization: it shows whether an apparent association is broadly supported across many configurations versus driven by a small number of points, and whether generator-specific clustering persists after residualization. Each panel plots \emph{metric residual} versus \emph{mAP residual} after regressing both on augmentation ratio (linear fit vs.\ \texttt{Aug\_Level}); dashed lines are ordinary least-squares fits in residual space (guide-to-the-eye only). Negative residual values indicate ``lower than expected given augmentation level'' rather than negative raw metrics. We focus on the from scratch regime because it is where residual signal is strongest in this benchmark; the analogous pretrained composites are omitted for brevity.

\section{Metric suite and implementation details}
\label{sec:extended-metrics}

This section summarizes the metric suite used in Step~2 and provides implementation details needed to reproduce the figures and analyses in the main paper. Definitions of all global (embedding-based) and object-centric metrics referenced throughout the paper are given in Appendix~\ref{app:metric_definitions}.

\subsection{Step 2 implementation details: labeling and matched-size bootstrap}
\label{app:step2_impl_details}

\paragraph{Teacher-assisted labeling and manual correction.}
We generate initial pseudo-labels on synthetic images using a fixed reference YOLOv11 teacher
(COCO-pretrained and fine-tuned on the corresponding real training split) with \texttt{conf}$=0.25$ and \texttt{iou}$=0.45$. We then manually review all teacher outputs and edit bounding boxes by removing false detections and adding missed instances. Synthetic images that do not contain the target class after manual inspection are removed. This workflow follows common human-in-the-loop ``pre-label + correction'' practice used to reduce full manual labeling cost while preserving label reliability~\citep{VEN2025110524, Ahmad2026SemiAutoAnnotation}.

\paragraph{Matched-size bootstrap for embedding-based metrics.}
Embedding-based distances (computed in both Inception-v3 and DINOv2 feature spaces) are estimated using a matched-size protocol to reduce the known sample-size sensitivity of Fr\'echet-style distances~\citep{Chong2020}. For each dataset, we fix the subset size to the $+10\%$ augmentation increment of the real training split ($N{=}179$ for Traffic Signs and COCO PottedPlant; $N{=}153$ for Cityscapes Pedestrian). We report the mean over $B{=}5$ bootstrap comparisons (and, in released tables, optionally the bootstrap dispersion).

Concretely, for each dataset we first sample $B{=}5$ real reference subsets $\{R^{(b)}\}_{b=1}^{B}$, each containing $N$ images drawn uniformly (without replacement) from the real training split. These real subsets are then \emph{kept fixed} and reused for \emph{all} generators, augmentation ratios, and embedding encoders, so that metric differences are attributable to the synthetic data rather than a changing real reference.

For a given dataset, generator, and augmentation ratio, let $S$ denote the available labeled synthetic pool at that ratio. For each bootstrap trial $b$, we form a size-matched synthetic subset $S^{(b)}$ of $N$ images drawn uniformly from $S$ (without replacement when $|S|\ge N$; otherwise with replacement). We then compute the embedding-based metric between the paired subsets $(R^{(b)}, S^{(b)})$ and average over $b=1,\dots,B$.

At the lowest augmentation level ($+10\%$), the synthetic pool size is $N$, so the synthetic draw is identical across bootstrap trials. At higher augmentation ratios (e.g., $+100\%$), the larger synthetic pool allows different $S^{(b)}$ draws across trials, yielding an average over multiple matched real--synthetic comparisons.

We use this matched-size design for two reasons. First, it avoids confounding metric values with the trivial effect of comparing different numbers of images, which can bias Fr\'echet-style estimates~\citep{Chong2020}. Second, it better reflects the effective synthetic data used at each augmentation ratio: computing metrics on the \emph{full} synthetic pool would produce values that are not comparable across ratios and would not correspond to the ratio-specific augmented training sets used in our YOLO experiments. We therefore interpret these metrics as \emph{ratio-conditional} quality/divergence estimates; they may differ (often be noisier and larger) than full-pool estimates computed with substantially more samples. For consistency, object-centric metrics are aggregated over the same $B$ trials.

\subsection{Metric definitions (global and object-centric)}
\label{app:metric_definitions}

This subsection defines the metrics reported throughout the paper and released in the accompanying result tables.
All \emph{global} metrics operate on an embedding function $\phi(\cdot)$ that maps an image to a feature vector.
In our experiments, $\phi$ is instantiated either by Inception-v3 or by DINOv2, but the metric definitions are unchanged; only the feature space differs.

\paragraph{Global (embedding-based) distribution metrics.}
Given real embeddings $\{\phi(x_i)\}_{i=1}^{n}$ and synthetic embeddings $\{\phi(\tilde{x}_j)\}_{j=1}^{m}$:

\begin{itemize}
    \item \textbf{Fr\'echet Distance (FID/FD; lower is better).}
    We compute the Fr\'echet distance between Gaussians fit to real and synthetic embeddings (mean/covariance form).
    When $\phi$ is Inception-v3 we refer to this as FID; when $\phi$ is DINOv2 we refer to this as FD. \citep{Heusel2017}

    \item \textbf{Effectively unbiased Fr\'echet distance (FID$_\infty$/FD$_\infty$; lower is better).}
    A bias-reduced/extrapolated variant intended to reduce finite-sample bias relative to standard FID/FD. \citep{Naeem2020}

    \item \textbf{Kernel Distance (KD; lower is better).}
    A kernel two-sample statistic (MMD-style) computed in embedding space, used as an alternative global distribution distance.

    \item \textbf{Precision and Recall (higher is better).}
    Measures of fidelity (precision) and diversity/coverage (recall) computed via neighborhood/manifold comparisons in feature space. \citep{kynkäänniemi2019improvedprecisionrecallmetric}

    \item \textbf{Density and Coverage (higher is better).}
    ``Improved'' variants related to precision/recall that quantify how densely generated samples populate the real manifold (density) and how much of the real manifold is covered (coverage). \citep{Naeem2020}

    \item \textbf{AuthPct / Authenticity percentage (higher is better).}
    A nearest-neighbor–style novelty/overfitting diagnostic intended to quantify how often synthetic samples appear non-memorized relative to the reference real set. \citep{Alaa2022}

    \item \textbf{SW approx / Sliced-Wasserstein approximation (lower is better).}
    An embedding-space sliced Wasserstein–style distance computed approximately via random projections. \citep{Deshpande_2018_CVPR,Wu_2019_CVPR}

    \item \textbf{CT and CT\_mod (higher is better; coverage-style).}
    Coverage-test style statistics that summarize how well the synthetic distribution covers the real distribution in feature space (with the ``mod'' variant using a modified normalization/estimator). \citep{meehan2020nonparametric}

    \item \textbf{FLS and FLS\_overfit (lower is better).}
    Feature Likelihood Score (and an overfit-oriented companion) that summarize how likely synthetic features are under a reference model fit to real features, acting as an embedding-space fit/overfitting diagnostic. \citep{c:jiralerspong2024feature}
\end{itemize}

\paragraph{Object-centric distribution metrics (computed from bounding-box statistics).}
In addition to global embedding-based metrics, we compute distances on simple label-derived statistics.
Let $n(I)$ be the number of boxes in image $I$, and let $c(I)$ be a per-image \emph{complexity score} (defined from box statistics; in our implementation it combines instance count with the prevalence of \emph{small/occluded} instances, as described in Section~\ref{sec:methodology}).

We report five object-centric metrics:
\begin{itemize}
\raggedright
\setlength{\itemsep}{2pt}
\item \textbf{Object-count Wasserstein}
(\texttt{object\_\allowbreak count\_\allowbreak wass\_\allowbreak mean}):
Wasserstein distance between the distributions of $n(I)$ for real vs.\ synthetic images.
\item \textbf{Object-count JSD}
(\texttt{object\_\allowbreak count\_\allowbreak jsd\_\allowbreak mean}):
Jensen--Shannon divergence between the distributions of $n(I)$ for real vs.\ synthetic images.
\item \textbf{Complexity Wasserstein}
(\texttt{complexity\_\allowbreak wass\_\allowbreak mean}):
Wasserstein distance between the distributions of $c(I)$ for real vs.\ synthetic images.
\item \textbf{Complexity JSD}
(\texttt{complexity\_\allowbreak jsd\_\allowbreak mean}):
Jensen--Shannon divergence between the distributions of $c(I)$ for real vs.\ synthetic images.
\item \textbf{Difficult-object ratio difference}
(\texttt{difficult\_\allowbreak object\_\allowbreak ratio\_\allowbreak diff\_\allowbreak mean}):
Absolute gap between real and synthetic fractions of ``difficult'' instances (here: \emph{small/occluded objects}), reported as a single scalar.

\end{itemize}

All reported global and object-centric metrics are the mean of $B$ bootstrap comparisons under the matched-size protocol described in Section~\ref{sec:methodology}, so that metric values are comparable across augmentation ratios.

\subsection{Correlation analysis details (residualization, BH--FDR, key-metric shortlist)}
\label{app:correlation_analysis_details}

\paragraph{Residualization (augmentation-controlled correlations).}
\phantomsection\label{app:residual_correlation_details}
For each dataset and regime, let $a$ denote augmentation ratio, $y$ denote held-out evaluation mAP, and $M$ denote a metric from Appendix~\ref{app:metric_definitions}. We fit the linear trends
\[
y(a) = \alpha + \beta a + \epsilon,
\qquad
M(a) = \gamma + \delta a + \eta,
\]
and compute correlations using: (i) raw pairs $(y, M)$, and (ii) residual pairs $(\epsilon, \eta)$. Residual correlations therefore quantify association \emph{beyond} the augmentation-level trend and directly target the fixed-budget generator-selection setting emphasized in the main paper. We report Spearman correlation (rank-based, robust to monotonic but non-linear relationships), with the appendix heatmaps focusing on residual Spearman for readability.

\paragraph{Multiple-testing correction (BH--FDR).}
\phantomsection\label{app:bh_fdr_details}
Because we test many metric--performance associations in parallel, naive significance testing would yield false positives by chance. We therefore control the false discovery rate using the Benjamini--Hochberg (BH) procedure~\citep{Benjamini1995}. For each training regime (from scratch vs.\ pretrained) and each correlation view (raw vs.\ residual/augmentation-controlled), we form the full matrix of per-cell correlation $p$-values across all metrics and all datasets shown in the corresponding full heatmap (i.e., all metric$\times$dataset cells with finite $p$-values). We then apply BH--FDR correction across these $p$-values to obtain a $q$-value for each cell, and mark cells with $q<0.05$ using an asterisk. To ensure consistent star annotations between the full-suite heatmaps and the compact key-metric heatmaps, $q$-values are computed once from the full $p$-value matrix and then subset to the rows shown in the key plots.

\paragraph{Key-metric shortlist for compact heatmaps (visualization only).}
\phantomsection\label{app:key_metric_selection_details}
The main paper includes compact ``key-metric'' heatmaps for readability (e.g., Figure~\ref{fig:heatmap_resid_fromscratch_key}). Key-metric selection affects only visualization (which rows are shown) and does not affect the underlying correlation computations. For a given regime, let $\rho_{\mathrm{resid}}(M,d)$ denote the residual Spearman correlation (augmentation-controlled) between metric $M$ and held-out test mAP for dataset $d$. We assign each metric a single strength score based on its strongest observed absolute association across datasets:
\[
s(M) \;=\; \max_{d} \left| \rho_{\mathrm{resid}}(M,d) \right|.
\]
We then select the top-$K$ metrics by $s(M)$ (with $K{=}15$ in the main paper) and order them by decreasing $s(M)$. All metrics defined in Appendix~\ref{app:metric_definitions} appear in the full residual heatmaps (Appendix~\ref{app:full_heatmaps}, Appendix~\ref{app:pretrained_heatmaps}).

\subsection{Full correlation heatmaps (complete metric suite, from scratch)}
\label{app:full_heatmaps}

This section reports full-suite Spearman correlation heatmaps between pre-training synthetic-data metrics (row; computed under the matched-size bootstrap protocol in Step~2) and held-out real test mAP@0.50:0.95 (column; Step~3), pooled over all generator--augmentation configurations within a dataset and regime. To support direct visual comparison, we show \emph{residual} correlations (controlling for augmentation ratio) side-by-side with \emph{raw} correlations (no augmentation control). Asterisks denote BH--FDR corrected significance at $q<0.05$, where BH--FDR is applied to the full set of metric$\times$dataset correlation tests in the corresponding heatmap (per regime and correlation view).

Residual correlations are computed by regressing both mAP and each metric on augmentation ratio within each dataset/regime and correlating the residuals. This targets the practical question: \emph{“At a fixed augmentation budget, do metric differences track generator-to-generator mAP differences?''}
 
\begin{figure*}[t]
    \centering
    \begin{subfigure}[t]{0.49\textwidth}
        \centering
        \includegraphics[width=\linewidth]{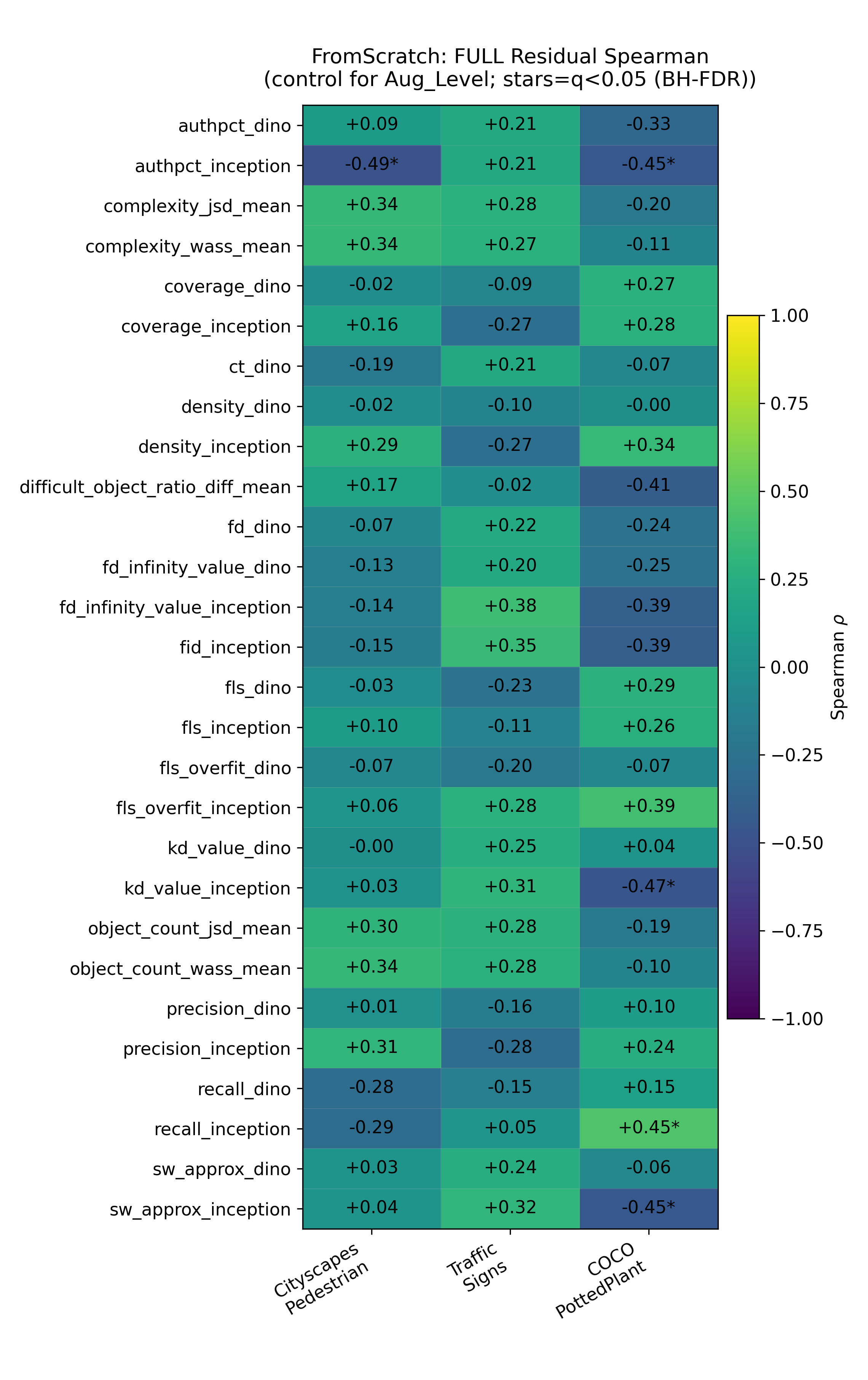}
        \caption{\small \textbf{Residual} Spearman (control for augmentation ratio).}
        \label{fig:heatmap_full_resid_spearman_fromscratch}
    \end{subfigure}\hfill
    \begin{subfigure}[t]{0.49\textwidth}
        \centering
        % raw PNG is narrower, so scale it down to match visual scale
        \includegraphics[width= \linewidth]{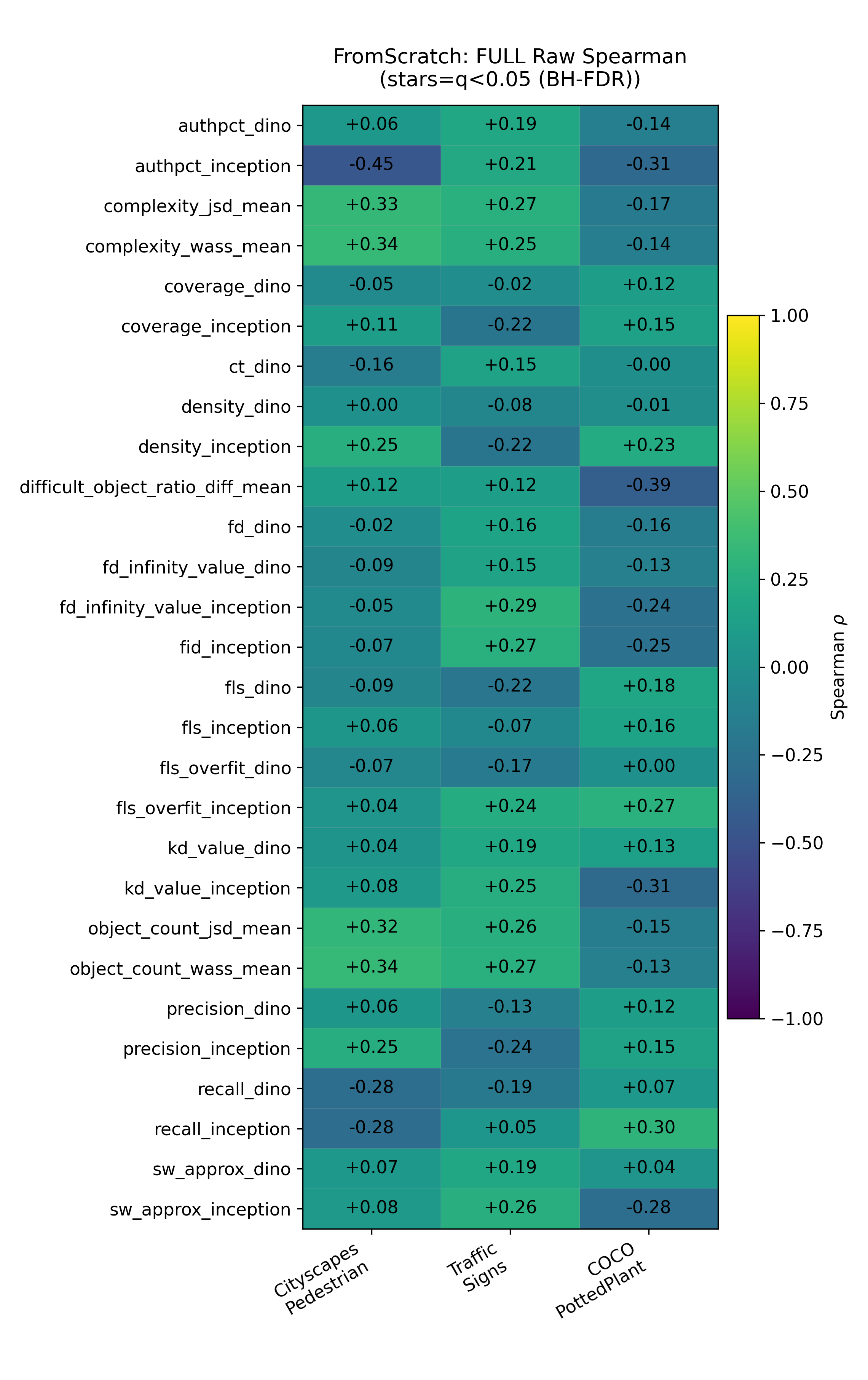}
        \caption{\small \textbf{Raw} Spearman (no augmentation control).}
        \label{fig:heatmap_full_raw_spearman_fromscratch}
    \end{subfigure}
    \caption{\small \textbf{From-Scratch:} Full-suite Spearman metric--mAP correlations. Left: residual Spearman correlations after controlling for augmentation ratio. Right: raw Spearman correlations (no augmentation control). Stars indicate BH--FDR significance at $q<0.05$.}
    \label{fig:heatmap_full_pair_fromscratch}
\end{figure*}

\paragraph{Interpretation (From-Scratch).}
The residual heatmap reinforces three main points from the Results section.
First, \textbf{metric--performance alignment is strongly dataset-dependent}:
both the sign and magnitude of residual correlations vary across the three regimes.
For example, several feature-space distance metrics (e.g., FID/FD/KD variants) show
moderate \emph{negative} residual association with mAP in COCO PottedPlant and Cityscapes Pedestrian
(consistent with the intuition that \emph{smaller} synthetic--real feature-space gaps can help),
while the same family can exhibit weaker or even sign-flipped behavior in the near-saturated Traffic Signs regime,
where residual mAP differences are small and generator ordering is less stable.

Second, after controlling for augmentation ratio and applying BH--FDR,
\textbf{only a small subset of metric--dataset pairs remains statistically significant}
(starred cells), indicating that many apparent associations in the raw view are largely explained by augmentation level
or do not robustly distinguish generators at a fixed augmentation ratio.

Third, object-centric distribution metrics (e.g., Wasserstein/JSD distances over object-count/complexity
and the difficult-object ratio difference) can show meaningful residual association in some regimes,
but \textbf{no single metric family is uniformly predictive across all datasets}.
Overall, the full-suite view supports the paper’s central conclusion that metric utility is regime-specific,
and that residual analysis is necessary to assess whether a metric provides signal beyond augmentation amount.

\paragraph{Raw vs.\ residual (From-Scratch).}
The raw heatmap provides a descriptive view of co-variation across the full generator--augmentation grid,
but it mixes two sources of variation: (i) the overall augmentation--performance trend and (ii) generator-to-generator
differences at a fixed augmentation budget. In this regime, none of the raw correlations are BH--FDR significant
(Figure~\ref{fig:heatmap_full_raw_spearman_fromscratch}). After controlling for augmentation ratio, a small number of
metric--dataset pairs become BH--FDR significant (Figure~\ref{fig:heatmap_full_resid_spearman_fromscratch}),
highlighting that augmentation control can materially change which associations remain detectable in the fixed-budget setting.

\subsection{Full correlation heatmaps (complete metric suite, pretrained)}
\label{app:pretrained_heatmaps}
\begin{figure*}[t]
    \centering
    \begin{subfigure}[t]{0.49\textwidth}
        \centering
        \includegraphics[width=\linewidth]{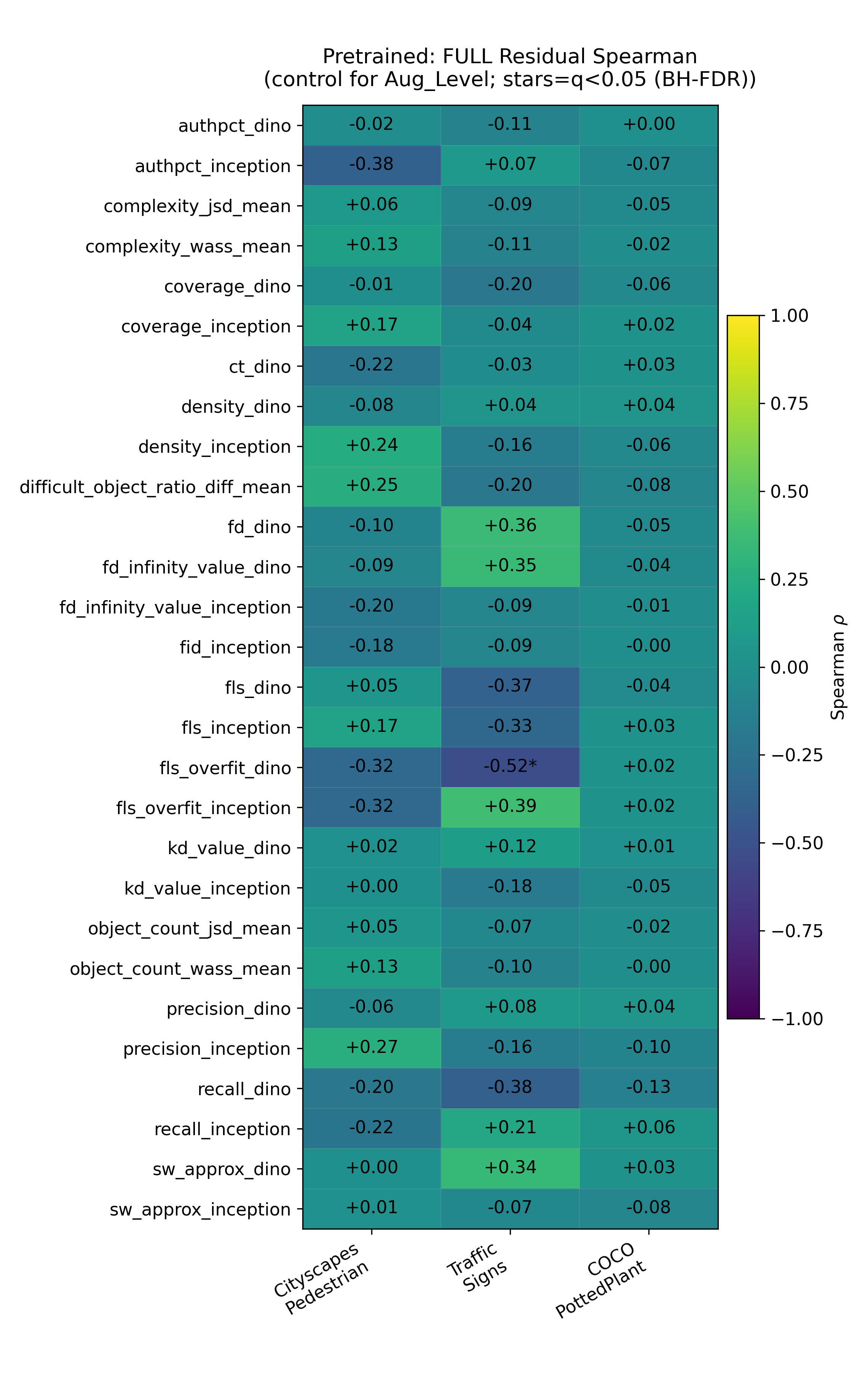}
        \caption{\small \textbf{Residual} Spearman (control for augmentation ratio).}
        \label{fig:heatmap_full_resid_spearman_pretrained}
    \end{subfigure}\hfill
    \begin{subfigure}[t]{0.49\textwidth}
        \centering
        \includegraphics[width= \linewidth]{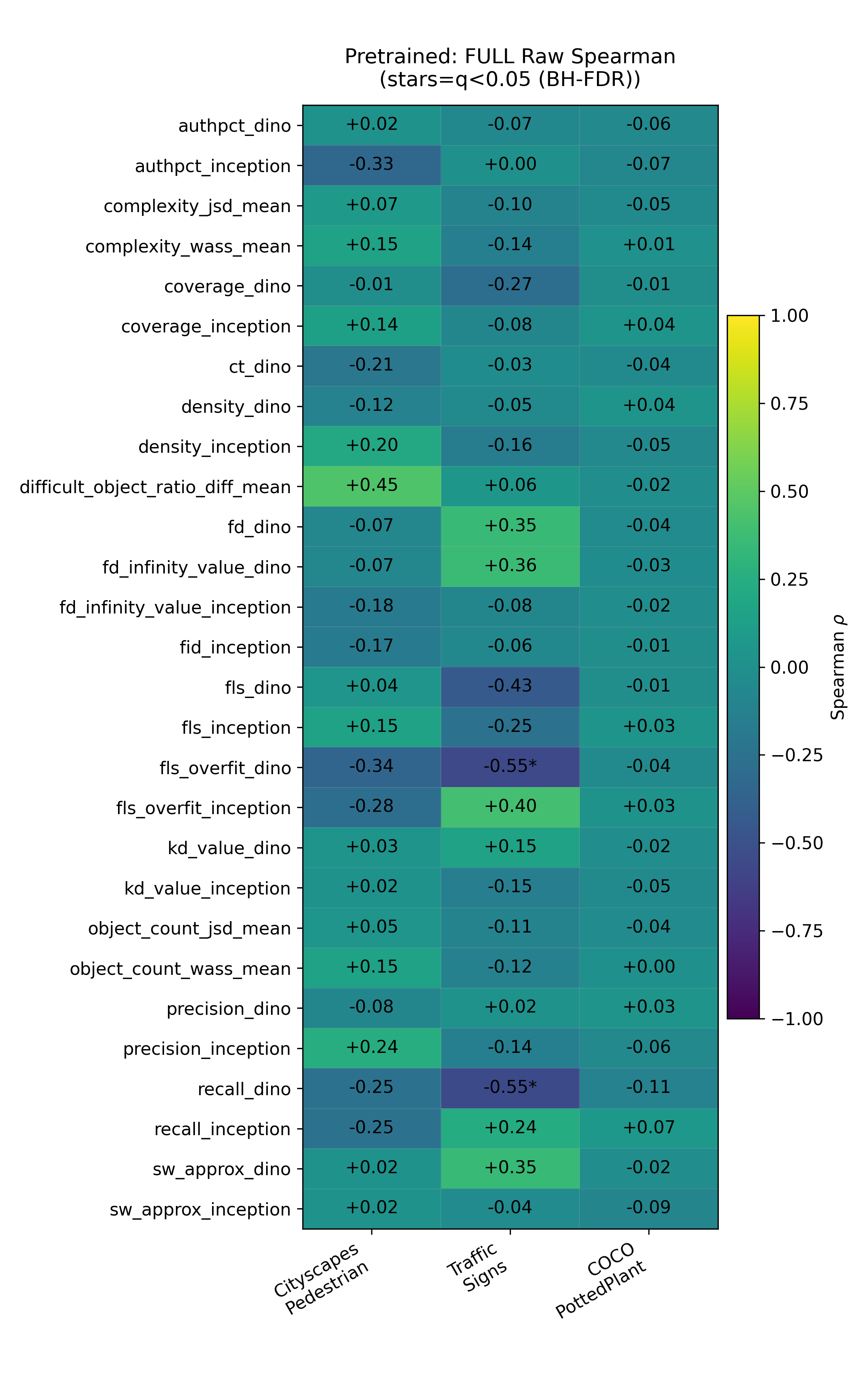}
        \caption{\small \textbf{Raw} Spearman (no augmentation control).}
        \label{fig:heatmap_full_raw_spearman_pretrained}
    \end{subfigure}
    \caption{\small \textbf{Pretrained:} Full-suite Spearman metric--mAP correlations. Left: residual Spearman correlations after controlling for augmentation ratio. Right: raw Spearman correlations (no augmentation control). Stars indicate BH--FDR significance at $q<0.05$.}
    \label{fig:heatmap_full_pair_pretrained}
\end{figure*}

\paragraph{Interpretation (Pretrained).}
In the COCO-pretrained regime, residual correlations are generally smaller in magnitude and rarely survive BH--FDR in the residual heatmap (Figure~\ref{fig:heatmap_full_resid_spearman_pretrained}),
with only an isolated significant association visible in this full-suite view (Traffic Signs: \texttt{fls\_overfit\_dino}, $\rho=-0.52$). This is consistent with the empirical behavior observed in the pretrained augmentation results: once starting from a strong pretrained representation, the remaining generator-to-generator differences (at a fixed augmentation ratio) are typically reduced and can be inconsistent across regimes, limiting the ability of generic global metrics or simple object-centric distribution metrics to reliably rank generators.
Practically, this suggests that \textbf{metric-based screening of generators is more plausible in the from scratch regime} (where residual mAP variation is larger) than under pretrained fine-tuning in these experiments.

\paragraph{Raw vs.\ residual (Pretrained).}
In the raw heatmap, BH--FDR-significant associations are limited to Traffic Signs and remain isolated (Figure~\ref{fig:heatmap_full_raw_spearman_pretrained}). After controlling for augmentation ratio, only \texttt{fls\_overfit\_dino} remains BH--FDR significant (Figure~\ref{fig:heatmap_full_resid_spearman_pretrained}),
while the other raw significant association(s) do not persist. Overall, this reflects weaker and less stable augmentation-controlled metric--mAP structure under COCO-pretrained fine-tuning in these experiments.

\subsection{Residualization robustness checks: per-budget correlations and categorical augmentation fixed effects}
\label{app:residualization_robustness}

The main analysis controls for augmentation amount via residualization with respect to augmentation ratio. Since augmentation–performance relationships can exhibit diminishing returns and other non-linearities, and each generator is evaluated at multiple augmentation levels (a repeated-measures design), we additionally report two complementary analyses that control augmentation without assuming a linear augmentation trend.

\paragraph{(R1) Per-augmentation-level correlations across generators.}
For each dataset and regime, and for each fixed budget $a \in \{25\%,50\%,100\%\}$, we compute Spearman correlation between metric values and held-out test mAP \emph{across generators only}. This removes augmentation as a confound by construction (augmentation is held fixed) and directly tests whether a metric tracks generator-to-generator differences at a fixed synthetic-data budget.

\paragraph{(R2) Categorical augmentation fixed-effects model.}

Pooling all augmentation levels, we fit a regression treating augmentation as a categorical fixed effect:
\[
\mathrm{mAP} = \alpha + \sum_{a}\theta_a\,\mathbf{1}[\mathrm{Aug{=}a}] + \beta_{\mathrm{FE}} M + \epsilon,
\]
and report the metric coefficient $\beta_{\mathrm{FE}}$ and its $p$-value. This controls for \emph{arbitrary} (non-linear) augmentation-level shifts via $\{\theta_a\}$, while estimating whether the metric explains additional variation in mAP beyond augmentation-as-category.

\paragraph{Handling undefined correlations.}
At some budgets a metric can be (near-)constant across generators (e.g., due to estimator saturation or limited variability), yielding undefined Spearman correlation (zero variance). We denote these cases with \textemdash and list the reason in the Notes column.

\begin{table*}[t]
\centering
\caption{\small Robustness checks that control augmentation without assuming a linear trend. $\rho_{25},\rho_{50},\rho_{100}$: within-budget Spearman correlations (across generators) at $a\in\{25\%,50\%,100\%\}$. $\beta_{\mathrm{FE}},p_{\mathrm{FE}}$: metric effect in a categorical augmentation fixed-effects regression. \textemdash denotes undefined Spearman correlation (zero metric variance; see Notes).}
\label{tab:residualization_robustness}

\begingroup
\setlength{\tabcolsep}{4pt}            % tighter columns (default is 6pt)
\renewcommand{\arraystretch}{1.05}     % slightly tighter rows (default 1.0)

\begin{adjustbox}{max width=0.98\textwidth}
\begin{tabular}{l l l r r r r r r l}
\toprule
\textbf{Regime} & \textbf{Dataset} & \textbf{Metric} &
$\boldsymbol{\rho_{25}}$ & $\boldsymbol{\rho_{50}}$ & $\boldsymbol{\rho_{100}}$ & $\boldsymbol{\bar{\rho}}$ &
$\boldsymbol{\beta_{\mathrm{FE}}}$ & $\boldsymbol{p_{\mathrm{FE}}}$ & \textbf{Notes} \\
\midrule
From-scratch & Pedestrian    & difficult\_object\_ratio\_diff\_mean ($\downarrow$) & \textemdash & \textemdash & +0.393 & +0.393 & +4.630 & 0.028 & 25:constant; 50:constant \\
From-scratch & PottedPlant   & difficult\_object\_ratio\_diff\_mean ($\downarrow$) & +0.143 & -0.714 & -0.429 & -0.333 & -0.324 & 0.001 & \textemdash \\
From-scratch & Traffic Signs & authpct\_inception ($\uparrow$)                      & +0.401 & +0.414 & +0.845 & +0.553 & +0.008 & 0.146 & \textemdash \\
Pretrained  & Pedestrian    & fls\_overfit\_inception ($\downarrow$)               & -0.086 & -0.314 & -0.200 & -0.200 & -0.000 & 0.072 & \textemdash \\
Pretrained  & PottedPlant   & fls\_overfit\_inception ($\downarrow$)               & -0.771 & +0.771 & -0.086 & -0.029 & +0.000 & 0.154 & \textemdash \\
Pretrained  & Traffic Signs & fls\_overfit\_dino ($\downarrow$)                    & -0.943 & -0.486 & -0.086 & -0.505 & -0.000 & <0.001 & \textemdash \\
\bottomrule
\end{tabular}
\end{adjustbox}

\endgroup
\end{table*}
\paragraph{Takeaway.}
These robustness checks target the same fixed-budget question as the residual Spearman heatmaps (Figure~\ref{fig:heatmap_resid_fromscratch_key} and Figure~\ref{fig:heatmap_full_resid_spearman_fromscratch}--\ref{fig:heatmap_full_resid_spearman_pretrained}) in Appendix, but control augmentation without assuming a linear augmentation--mAP trend. The within-budget correlations $(\rho_{25},\rho_{50},\rho_{100})$ test whether a metric tracks \emph{generator-to-generator} mAP differences at fixed synthetic budgets, while the categorical fixed-effects coefficient $\beta_{\mathrm{FE}}$ tests whether the metric explains additional variation after absorbing augmentation-level shifts via $\{\theta_a\}$. In the from scratch regime, the shortlisted metrics show non-trivial within-budget associations (notably in PottedPlant and Traffic Signs), and the fixed-effects model yields statistically detectable pooled effects in Pedestrian and PottedPlant (Table~\ref{tab:residualization_robustness}); Traffic Signs is not significant under the fixed-effects test ($p_{\mathrm{FE}}=0.146$). In Pedestrian (from scratch), $\rho_{25}$ and $\rho_{50}$ are undefined because the metric is constant across generators at those budgets (Notes), so only $\rho_{100}$ provides a ranking-based check, whereas $\beta_{\mathrm{FE}}$ still summarizes the pooled association. In the pretrained regime, within-budget correlations are generally smaller and/or vary across budgets, and most $\beta_{\mathrm{FE}}$ estimates are not statistically distinguishable from zero; the main exception is Traffic Signs, where \texttt{fls\_overfit\_dino} has $p_{\mathrm{FE}}<0.001$ and the residual heatmap shows the only BH--FDR-significant cell (Appendix Figure~\ref{fig:heatmap_full_resid_spearman_pretrained}).

\subsection{Decision-oriented generator screening at fixed budgets}
\label{app:decision_screening}

Sections~\ref{sec:results-metric-alignment} quantify \emph{association} between pre-training metrics and downstream mAP. However, practitioners often need a \emph{decision rule}: given a fixed synthetic-data budget, which generator should be chosen \emph{before} training YOLO? We therefore evaluate a simple metric-based screening protocol at fixed augmentation budgets $a \in \{25\%,50\%,100\%\}$. Table~\ref{tab:decision_screening_summary} summarizes the decision-oriented screening results
(averaged over budgets 25\%, 50\%, and 100\%).

\paragraph{Protocol.}
For each dataset and initialization regime, and for each fixed budget $a$, we (i) collect the held-out test mAP for each generator at budget $a$, (ii) rank generators by a candidate metric using its standard direction (e.g., lower distance/divergence is better; higher coverage/precision is better), and (iii) select the top-ranked generator according to the metric.

\paragraph{Decision metrics.}
We report three standard decision-oriented summaries:
(1) \textbf{Kendall rank correlation} $\tau \in [-1,1]$ between the metric-induced generator ranking and the mAP ranking at budget $a$ (higher is better; $\tau{=}1$ indicates identical rankings)~\citep{kendall1938new};
(2) \textbf{Top-1 accuracy} (whether the metric selects the best-mAP generator at budget $a$);
and (3) \textbf{regret} (mAP gap to the best generator),
\[
\mathrm{Regret}(a) \;=\; \max_{g}\,\mathrm{mAP}(g,a)\;-\;\mathrm{mAP}(\hat{g},a),
\]
where $\hat{g}$ is the generator selected by the metric (lower is better; regret $=0$ is optimal). Regret is a standard way to evaluate selection rules under a fixed budget~\citep{bubeck2012regret}.
We summarize results by averaging over budgets $\{25\%,50\%,100\%\}$.

\begin{table}[!t]
\centering
\caption{\small Decision-oriented screening summary (averaged over budgets 25\%, 50\%, and 100\%). Per dataset/regime we report the \emph{single} shortlisted metric with the lowest average regret. Arrows indicate the ranking direction used for screening ($\downarrow$ = lower is better, $\uparrow$ = higher is better).}
\label{tab:decision_screening_summary}
\footnotesize
\setlength{\tabcolsep}{3pt}
\renewcommand{\arraystretch}{1.05}

\resizebox{\columnwidth}{!}{%
\begin{tabular}{l l l c c c}
\toprule
\textbf{Regime} & \textbf{Dataset} & \textbf{Best metric (dir.)} & \textbf{Avg $\tau$} & \textbf{Top-1} & \textbf{Avg regret} \\
\midrule
From-scratch & Pedestrian   & \texttt{fls\_overfit\_inception} ($\downarrow$) & 0.156 & 0/3 & 0.0028 \\
From-scratch & PottedPlant  & \texttt{kd\_value\_inception} ($\downarrow$)     & 0.422 & 3/3 & 0.0000 \\
From-scratch & Traffic Signs& \texttt{authpct\_inception} ($\uparrow$)         & 0.467 & 1/3 & 0.0079 \\
\midrule
Pretrained   & Pedestrian   & \texttt{fls\_overfit\_inception} ($\downarrow$) & 0.200 & 1/3 & 0.0045 \\
Pretrained   & PottedPlant  & \texttt{fd\_dino} ($\downarrow$)                & -0.022 & 1/3 & 0.0052 \\
Pretrained   & Traffic Signs& \texttt{fls\_overfit\_dino} ($\downarrow$)      & 0.467 & 1/3 & 0.0001 \\
\bottomrule
\end{tabular}}
\end{table}

\paragraph{Heatmap-shortlist (non-cherry-picked).}
To avoid testing a large number of metrics, we restrict to a shortlist derived directly from the residual heatmaps in this paper: we include metrics that either (i) achieve BH--FDR significance ($q<0.05$) in any residual Spearman cell, or (ii) satisfy $\max_{\text{dataset}} |\rho_{\text{resid}}| \ge 0.35$. This links the screening study directly to the metrics that exhibit signal after controlling for augmentation ratio.

\paragraph{Interpretation and relation to the residual heatmaps.}
This decision-style view remains consistent with the paper’s central message that \emph{useful screening is regime-dependent}. In the \textbf{from scratch} regime—where generator differences are typically larger—the best-performing screening metrics align with the metric families that exhibit residual signal in the heatmaps. In particular, \texttt{kd\_value\_inception} is the strongest practical screening rule on \textbf{PottedPlant}, showing clear rank agreement (Avg $\tau{=}0.422$), perfect Top-1 selection (3/3 budgets), and zero average regret; it selects the best-mAP generator at all three evaluated budgets in this setting. On \textbf{Traffic Signs}, \texttt{authpct\_inception} remains the most competitive screening metric: it achieves moderate average ranking agreement (Avg $\tau{=}0.467$) but only occasional Top-1 success (1/3), consistent with a near-saturated regime where small mAP gaps make consistent selection difficult despite detectable association. On \textbf{Pedestrian} (from scratch), screening is weaker: the best metric (\texttt{fls\_overfit\_inception}) yields only small average agreement (Avg $\tau{=}0.156$) and fails to pick the best generator (0/3), though the regret remains modest, suggesting either limited separation between top generators at the evaluated budgets or instability in the identity of the best generator across budgets.
In the \textbf{pretrained} regime, average regret is generally small—especially on \textbf{Traffic Signs}, where the best metric (\texttt{fls\_overfit\_dino}) achieves very low regret—reflecting limited headroom across generators once initialization is strong. Accordingly, screening provides weaker and less consistent ranking agreement (e.g., near-zero Avg $\tau$ on \textbf{PottedPlant} for \texttt{fd\_dino}), indicating that metric-based screening is less reliable as a ranking surrogate in this regime even when it often selects near-best generators in absolute mAP. We treat these screening results as exploratory within this benchmark; a natural next step is to learn a composite predictor (e.g., regression over the metric suite) and evaluate transfer to additional datasets.

\end{document}